\documentclass[sigconf]{acmart}

\usepackage{subfigure}
\usepackage{multirow}
\usepackage{xspace}
\usepackage{listings}
\usepackage{xcolor}
\usepackage{makecell}
\usepackage{CJKutf8}
\usepackage{bm}

\newcommand{\blue}[1]{\textcolor{blue}{#1}}

\newcommand{\para}[1]{{\vspace{2pt} \bf \noindent #1 \hspace{1pt}}}

\AtBeginDocument{%
  \providecommand\BibTeX{{%
    \normalfont B\kern-0.5em{\scshape i\kern-0.25em b}\kern-0.8em\TeX}}}

\copyrightyear{2022}
\acmYear{2022}
\setcopyright{acmcopyright}\acmConference[KDD '22]{Proceedings of the 28th ACM SIGKDD Conference on Knowledge Discovery and Data Mining}{August 14--18, 2022}{Washington, DC, USA}
\acmBooktitle{Proceedings of the 28th ACM SIGKDD Conference on Knowledge Discovery and Data Mining (KDD '22), August 14--18, 2022, Washington, DC, USA}
\acmPrice{15.00}
\acmDOI{10.1145/3534678.3539374}
\acmISBN{978-1-4503-9385-0/22/08}

\begin{document}
\fancyhead{}
\title{Model Degradation Hinders Deep Graph Neural Networks}
\author{Wentao Zhang$^\dagger$, Zeang Sheng$^{\dagger}$, Ziqi Yin$^{\ddagger}$, Yuezihan Jiang$^{\dagger}$, \\Yikuan Xia$^{\dagger}$, Jun Gao$^{\dagger}$,
Zhi Yang$^{\dagger \S}$, 
Bin Cui$^{\dagger \S}$}
\affiliation{
 {{$^\dagger$}{School of CS \& Key Laboratory of High Confidence Software Technologies, Peking University}}~~~~~\\
 {{$^{\S}$}{Center for Data Science, Peking University \& National Engineering Laboratory for Big Data Analysis and Applications}}~~~~~\\
 $\ddagger$Beijing Institute of Technology $^\ddagger$\{ziqiyin18\}@bit.edu.cn\country{}}
\affiliation{
$^\dagger$\{wentao.zhang, shengzeang18, jiangyuezihan, 2101111522, gaojun, yangzhi, bin.cui\}@pku.edu.cn\country{}
}
\renewcommand{\shortauthors}{Wentao Zhang, et al.}

\begin{abstract}
  
Graph Neural Networks (GNNs) have achieved great success in various graph mining tasks.
However, drastic performance degradation is always observed when a GNN is stacked with many layers.
As a result, most GNNs only have shallow architectures, which limits their expressive power and exploitation of deep neighborhoods.
Most recent studies attribute the performance degradation of deep GNNs to the \textit{over-smoothing} issue.
In this paper, we disentangle the conventional graph convolution operation into two independent operations: \textit{Propagation} (\textbf{P}) and \textit{Transformation} (\textbf{T}).
Following this, the depth of a GNN can be split into the propagation depth ($D_p$) and the transformation depth ($D_t$).
Through extensive experiments, we find that the major cause for the performance degradation of deep GNNs is the \textit{model degradation} issue caused by large $D_t$ rather than the \textit{over-smoothing} issue mainly caused by large $D_p$.
Further, we present \textit{Adaptive Initial Residual} (AIR), a plug-and-play module compatible with all kinds of GNN architectures, to alleviate the \textit{model degradation} issue and the \textit{over-smoothing} issue simultaneously.
Experimental results on six real-world datasets demonstrate that GNNs equipped with AIR outperform most GNNs with shallow architectures owing to the benefits of both large $D_p$ and $D_t$, while the time costs associated with AIR can be ignored.

\end{abstract}




\maketitle

{\fontsize{8pt}{8pt} \selectfont
\textbf{ACM Reference Format:}\\
Wentao Zhang, Zeang Sheng, Ziqi Yin, Yuezihan Jiang, Yikuan Xia, Jun Gao, Zhi Yang, Bin Cui. 2022. Model Degradation Hinders Deep Graph Neural Networks. In \textit{Proceedings of the 28th ACM SIGKDD Conference on Knowledge Discovery and Data Mining (KDD '22), August 14--18, 2022, Washington, DC, USA} ACM, New York, NY, USA, 11 pages. https://doi.org/10.1145/3534678.3539374}

\section{Introduction}
The recent success of Graph Neural Networks (GNNs)~\cite{zhang2020deep} has boosted research on various data mining and knowledge discovery tasks on graph-structured data. 
GNNs provide a universal framework to tackle node-level, edge-level, and graph-level tasks, including social network analysis~\cite{qiu2018deepinf}, chemistry and biology~\cite{DBLP:conf/nips/DaiLCDS19}, recommendation~\cite{wu2020graph, jiang2022zoomer}, natural language processing~\cite{bastings2017graph}, and computer vision~\cite{qi2018learning}. 


In recent years, the graph convolution operation proposed by Graph Convolutional Network (GCN)~\cite{kipf2016semi} gradually becomes the canonical form of layer designs in most GNN models~\cite{wu2019simplifying,klicpera2018predict,zhang2021rod}.
Specifically, the graph convolution operation in GCN can be disentangled into two independent operations: \textit{Propagation} (\textbf{P}) and \textit{Transformation} (\textbf{T}).
The \textbf{P} operation can be viewed as a particular form of the Laplacian smoothing~\cite{DBLP:journals/corr/abs-1905-09550},
after which the representations of nearby nodes would become similar.
The \textbf{P} operation greatly reduces the difficulties of the downstream tasks since most real-world graphs follow the homophily assumption~\cite{mcpherson2001birds} that connected nodes tend to belong to similar classes.
The \textbf{T} operation applies non-linear transformations to the node representations, thus enabling the model to capture the data distribution of the training samples.
After the disentanglement, the depth of a GNN is split into the propagation depth ($D_p$) and the transformation depth ($D_t$).
A GNN with larger $D_p$ enables each node to exploit information from deeper neighborhoods, and a larger $D_t$ gives the model higher expressive power.

Despite the remarkable success of GNNs, deep GNNs are rarely applied in various tasks as simply stacking many graph convolution operations leads to drastic performance degradation~\cite{kipf2016semi}.
As a result, most GNNs today only have shallow architectures~\cite{kipf2016semi, wu2019simplifying, DBLP:conf/iclr/VelickovicCCRLB18}, which limits their performance.
Many novel architectures and strategies have been proposed to alleviate this problem, yet they disagree on the major cause for the performance degradation of deep GNNs. 
Among the suggested reasons, most existing studies~\cite{feng2020graph,chen2020measuring,DBLP:conf/iclr/ZhaoA20, godwin2021very, rong2019dropedge, miao2021lasagne, DBLP:conf/iclr/ChienP0M21, yan2021two, cai2020note} consider the \textit{over-smoothing} issue as the major cause for the performance degradation of deep GNNs.
The \textit{over-smoothing} issue~\cite{li2018deeper} refers to the phenomenon that node representations become indistinguishable after many graph convolution operations.
It is proved in~\cite{zhang2021node} that the differences between the node representations are only determined by the node degrees after applying infinity \textbf{P} operations.

In this paper, we conduct a comprehensive analysis to review the \textit{over-smoothing} issue in deep GNNs and try to identify the major cause for the performance degradation of deep GNNs.
We find that the \textit{over-smoothing} issue does happen after dozens of \textbf{P} operations, but the performance degradation of deep GNNs is observed far earlier than the appearance of the \textit{over-smoothing} issue.
Thus, the \textit{over-smoothing} issue is not the major cause for the performance degradation of deep GNNs.

On the contrary, the experiment results illustrate that the major cause for the performance decline is the \textit{model degradation} issue caused by large $D_t$ (i.e., stacking many \textbf{T} operations).
The \textit{model degradation} issue has been known to the community since the discussion in~\cite{he2016deep}.
It refers to the phenomenon that both the training accuracy and the test accuracy drop when the layer number of the network increases.
Although both are caused by increasing the layer number, the \textit{model degradation} issue is different from the \textit{overfitting} issue since the training accuracy remains high in the latter.
There have been many studies explaining the causes for the appearance of the \textit{model degradation} issue in deep neural networks~\cite{he2016deep, balduzzi2017shattered}.

To help GNNs to enjoy the benefits of both large $D_p$ and $D_t$, we propose \textit{Adaptive Initial Residual} (AIR), a plug-and-play module that can be easily combined with all kinds of GNN architectures.
Adaptive skip connections are introduced among the \textbf{P} and the \textbf{T} operations by AIR which alleviates the \textit{model degradation} issue and the \textit{over-smoothing} issue at the same time.
Experiment results on six real-world datasets show that simple GNN methods equipped with AIR outperform most GNNs that only have shallow architectures.
Further, simple GNN methods equipped with AIR show better or at least steady predictive accuracy as $D_p$ and $D_t$ increases, which validates the positive effects of AIR on fighting against the \textit{model degradation} issue and the \textit{over-smoothing} issue.




\begin{figure}[tpb!]
	\centering
	\includegraphics[width=.95\linewidth]{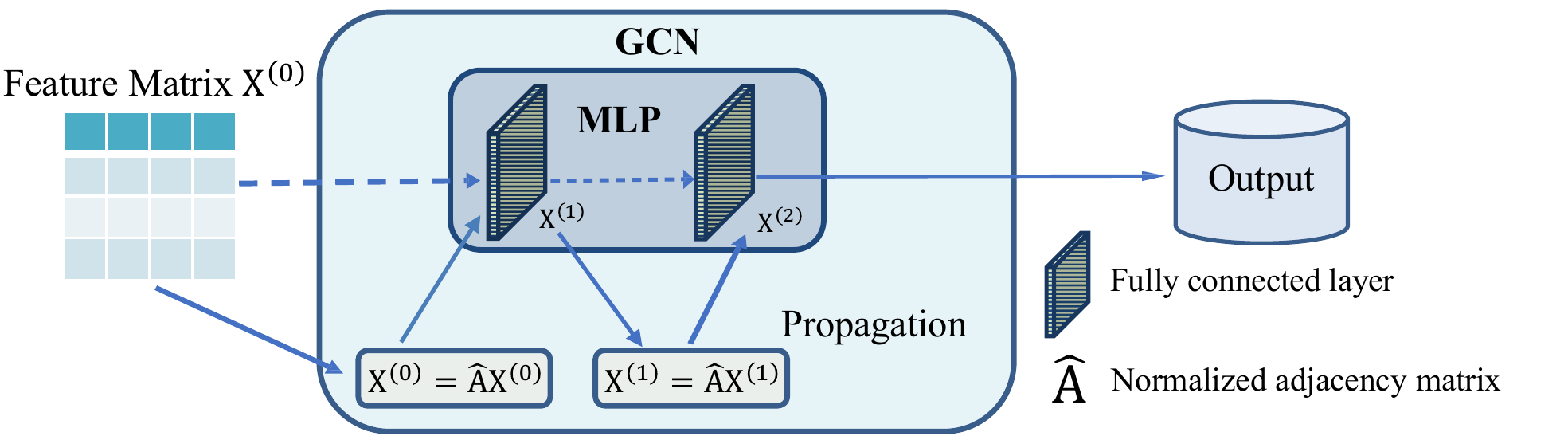}
 	\vspace{-2mm}
	\caption{The relationship between GCN and MLP.}
	\label{fig:GCN}
	\vspace{-2mm}
\end{figure}

\section{Preliminary}
In this section, we first explain the problem formulation.
Then we disentangle the graph convolution operation in GCN into two independent operations: \textit{Propagation} (\textbf{P}) and \textit{Transformation} (\textbf{T}).
This disentanglement split the depth of a GNN model into the propagation depth ($D_p$) and the transformation depth ($D_t$).
After that, the theoretical benefits of enlarging $D_p$ and $D_t$ will be discussed briefly.
Finally, we classify the existing GNN architectures into three categories according to the ordering of the \textbf{P} and \textbf{T} operations.

\subsection{Problem Formalization}
In this paper, we consider an undirected graph $\mathcal{G}$ = ($\mathcal{V}$, $\mathcal{E}$) with $|\mathcal{V}| = N$ nodes and $|\mathcal{E}| = M$ edges. 
$\mathbf{A}$ is the adjacency matrix of $\mathcal{G}$, weighted or not. 
Each node possibly has a feature vector $\in \mathbb{R}^d$, stacking up to an $N \times d$ matrix $\mathbf{X}$.
And $\mathbf{X}_i$ refers to the feature vector of node $i$.
$\mathbf{D}=\operatorname{diag}\left(d_{1}, d_{2}, \cdots, d_{n}\right) \in \mathbb{R}^{N \times N}$ denotes the degree matrix of $\mathbf{A}$, where $d_{i}=\sum_{j \in \mathcal{V}} \mathbf{A}_{i j}$ is the degree of node $i$. 
In this paper, we focus on the semi-supervised node classification task, where only part of the nodes in $\mathcal{V}$ are labeled.
$\mathcal{V}_l$ denotes the labeled node set, and $\mathcal{V}_u$ denotes the unlabeled node set.
The goal of this task is to predict the labels for nodes in $\mathcal{V}_u$ under the limited supervision of labels for nodes in $\mathcal{V}_l$.

\subsection{Graph Convolution Operation}
Following the classic studies in the graph signal processing field~\cite{sandryhaila2013discrete}, the graph convolution operation is first proposed in~\cite{bruna2013spectral}.
However, the excessive computation cost of eigendecomposition hinders~\cite{bruna2013spectral} from real-world practice.
Graph Convolutional Network (GCN)~\cite{kipf2016semi} proposes a much-simplified version of previous convolution operations on graph-structured data, making it the first feasible attempt on GNNs.
In recent years, the graph convolution operation proposed by GCN has gradually become the canonical form of most GNN architectures~\cite{zhang2020reliable,miao2021degnn,zhang2021rod,yang2020factorizable,yang2020distilling}. 

The graph convolution operation in GCN is formulated as:
\begin{equation}
\small
    \text{Graph Convolution}(\mathbf{X})=\sigma\big(\mathbf{\hat{A}}\mathbf{X}\mathbf{W}), \quad \mathbf{\hat{A}} = \widetilde{\mathbf{D}}^{-\frac{1}{2}}\widetilde{\mathbf{A}}\widetilde{\mathbf{D}}^{-\frac{1}{2}},
    \label{eq_GC}
\end{equation}

\noindent where $\mathbf{\hat{A}}$ is the normalized adjacency matrix, $\widetilde{\mathbf{A}}=\mathbf{A}+\mathbf{I}_{N}$ is the adjacency matrix $\mathbf{A}$ with self loops added, and $\mathbf{I}_{N}$ is the identity matrix.
$\mathbf{\hat{D}}$ is the corresponding degree matrix of $\mathbf{\hat{A}}$.
$\mathbf{W}$ is the learnable transformation matrix and $\sigma$ is the non-linear activation function.

From an intuitive view, the graph convolution operation in GCN firstly propagates the representation of each node to their neighborhoods and then transforms the propagated representations to specific dimensions by non-linear transformation.
"Graph convolution operation in GCN" is referred to as "graph convolution operation" in the rest of this paper if not specified.

\subsection{\textit{Propagation} (\textbf{P}) and \textit{Transformation} (\textbf{T}) Operations}
From the intuitive view in the above subsection, the graph convolution operation can be disentangled into two consecutive operations realizing different functionalities: \textit{Propagation} (\textbf{P}) and \textit{Transformation} (\textbf{T}).
Their corresponding formula form is as follows:
\begin{align}
    \label{eq_EP}
	\text{\textit{Propagation}}(\mathbf{X}) &= \textbf{P}(\mathbf{X}) = \mathbf{\hat{A}}\mathbf{X}, \\
	\label{eq_ET}
	\text{\textit{Transformation}}(\mathbf{X}) &= \textbf{T}(\mathbf{X})= \sigma(\mathbf{X}\mathbf{W}),
\end{align}
\noindent where $\mathbf{\hat{A}}$, $\mathbf{W}$ and $\sigma$ all have the same meanings as in Equation~\ref{eq_GC}.
It is evident that conducting the graph convolution operation is equivalent to first conducting the \textbf{P} operation then conducting the \textbf{T} operation, which can be expressed as follows:
\begin{equation}
\small
    \text{Graph Convolution}(\mathbf{X}) = \textbf{T}(\textbf{P}(\mathbf{X})). \nonumber
\end{equation}

GCN defines its model depth as the number of graph convolution operations in the model since it considers one graph convolution operation as one layer.
However, after the disentanglement, we can describe the depths of GNNs more precisely by two new metrics: the propagation depth ($D_p$) and the transformation depth ($D_t$).

Figure~\ref{fig:GCN} shows an illustrative example of a two-layer GCN.
To note that, the GCN will degrade to an MLP if the normalized adjacency matrix $\mathbf{\hat{A}}$ is set to the identity matrix $\mathbf{I}_N$, i.e., removing all the \textbf{P} operations in the model.

\subsection{Theoretical Benefits of Deep GNNs}
GCN achieves the best performance when composed of only two or three layers.
There have been many studies aiming at designing deep GNNs recently, and some of them~\cite{liu2020towards, zhu2021simple} achieve state-of-the-art performance on various tasks.
In this subsection, we will briefly discuss the theoretical benefits of deep GNNs and explain that both enlarging $D_p$ and $D_t$ will increase the model expressivity.

\subsubsection{Benefits of enlarging $D_p$}
Enlarging $D_p$ is equivalent to enlarging the receptive field of each node.
It is proved in~\cite{morris2019weisfeiler} that GCN has the same expressive power as the first-order Weisfeiler-Lehman graph isomorphism test.
Thus, enlarging the receptive field of each node makes it easier for the model to discriminate between two different nodes since it is more probable that they have highly different receptive fields.
~\cite{cong2021provable} proves that once the model is properly trained, the expressive power of GCN grows strictly as the layer number increases due to the enlargement of the receptive field.
To sum up, enlarging $D_p$ increases the model expressivity, which can be proved in the view of the Weisfeiler-Lehman test.

\subsubsection{Benefits of enlarging $D_t$}
Analyzing the change of the model expressivity when enlarging $D_t$ is easier than when enlarging $D_p$.
All know that the expressive power of a Multi-Layer Perceptron (MLP) grows strictly along with the increase of the layer number.
As introduced in the previous subsection, $D_t$ stands for the number of the non-linear transformations contained in the model.
Thus, enlarging $D_t$, i.e., increasing the number of non-linear transformations, also increases the model expressivity.

\subsection{Three Categories of GNN Architectures}
According to the ordering the model arranges the \textbf{P} and \textbf{T} operations, we roughly classify the existing GNN architectures into three categories: \textbf{PTPT}, \textbf{PPTT}, and \textbf{TTPP}.

\subsubsection{\textbf{PTPT}}
\textbf{PTPT} architecture is the original GNN design that is proposed by GCN~\cite{kipf2016semi}, and is widely adopted by mainstream GNNs, like GraphSAGE~\cite{hamilton2017inductive}, GAT~\cite{DBLP:conf/iclr/VelickovicCCRLB18}, and GraphSAINT~\cite{DBLP:conf/iclr/ZengZSKP20}.
\textbf{PTPT} architecture still uses the graph convolution operation in GCN, where the \textbf{P} and \textbf{T} operations are entangled.
In a more general view, the \textbf{P} and \textbf{T} operations are organized in a order like \textbf{PTPT}...\textbf{PT} in the \textbf{PTPT} architecture.
As a result, \textbf{PTPT} architecture has the strict restriction that $D_p = D_t$.
For example, if a model wants to enlarge the receptive field of each node, the natural idea is to enlarge $D_p$.
However, enlarging $D_p$ in the \textbf{PTPT} architecture requires enlarging $D_t$ at the same time. 
It would add a significant number of training parameters to the model, which exacerbates training difficulty.

\subsubsection{\textbf{PPTT}}
\textbf{PPTT} architecture is first proposed by SGC~\cite{wu2019simplifying}, which claims that the strength of GNN lies mainly in not the \textbf{T} operation but the \textbf{P} operation.
It disentangles the graph convolution operation and presents the \textbf{PPTT} architecture, where the \textbf{P} and \textbf{T} operations are arranged as \textbf{PP}...\textbf{PTT}...\textbf{T}.
This architecture is then adopted by many recent GNN studies~\cite{zhang2022pasca}, e.g., GAMLP~\cite{zhang2021gamlp}, SIGN~\cite{frasca2020sign}, S$^2$GC~\cite{zhu2021simple}, and GBP~\cite{DBLP:conf/nips/ChenWDL00W20}.
Compared with \textbf{PTPT} architecture, \textbf{PPTT} architecture breaks the chain of $D_p = D_t$, thus has more flexible design options.
For the same scenario that a model wants to enlarge the receptive field of each node, \textbf{PPTT} architecture can add the number of the stacked \textbf{P} operations (i.e., enlarging $D_p$) without changing $D_t$, which avoids increasing the training difficulty.
Besides, the \textbf{P} operations are only needed to be executed once during preprocessing since they can be fully disentangled from the training process.
This valuable property of \textbf{PPTT} architecture enables it with high scalability and efficiency.

\subsubsection{\textbf{TTPP}}
\textbf{TTPP} is another disentangled GNN architecture, which was first proposed by APPNP~\cite{klicpera2018predict}.
Being the dual equivalent to the \textbf{PPTT} architecture, \textbf{TTPP} architecture orders the \textbf{P} and \textbf{T} operations as \textbf{TT}...\textbf{TPP}...\textbf{P}, where the behavior of the stacked \textbf{P} operations can be considered as label propagation.
DAGNN~\cite{liu2020towards}, AP-GCN~\cite{spinelli2020adaptive}, GPR-GNN~\cite{DBLP:conf/iclr/ChienP0M21} and many other GNN models all follow the \textbf{TTPP} architecture.
Although \textbf{TTPP} architecture also enjoy the flexibility brought by the disentanglement of the graph convolution operation, it is much less scalable than \textbf{PPTT} architecture as the stacked \textbf{P} operations are entangled with the training process, which hinders its application on large graphs.
On the positive side, the stacked \textbf{T} operations can considerably reduce the dimensionality of the input features, which boosts the efficiency of \textbf{TTPP} architecture in the later stacked \textbf{P} operations.

\section{Empirical Analysis of the \textit{Over-smoothing} Issue}
In this section, we first define \textit{smoothness level} and introduce metrics to measure it at node and graph levels.
Then, we review the \textit{over-smoothing} issue and the reasons why it happens.
The rest of this section is an empirical analysis trying to figure out whether the \textit{over-smoothing} issue is the major cause behind the performance degradation of deep GNNs.

\subsection{Smoothness Measurement}
\label{sec.smooth_metric}
Smoothness level measures the similarities among node pairs in the graph.
Concretely, a higher smoothness level indicates that it happens with a higher probability that two randomly picked nodes from the given node set are similar.

Here we borrow the metrics from DAGNN~\cite{liu2020towards} to evaluate the smoothness level both at the node level and graph level.
However, we replace the Euclidean distance in~\cite{liu2020towards} with the cosine similarity to better measure the similarity between two papers in the citation network since the features of nodes are always constructed by word frequencies.
We formally define ``Node Smoothness Level (NSL)'' and ``Graph Smoothness Level (GSL)'' as follows:
\begin{definition}[\textbf{Node Smoothing Level}]
\label{df.nsl}
The Node Smoothing Level of node $i$, $NSL_i$, is defined as: 
\begin{equation}
\small
NSL_i = \frac{1}{N-1}\sum_{j\in \mathcal{V}, j\neq i}\frac{\mathbf{X}_i \cdot \mathbf{X}_j}{|\mathbf{X}_i||\mathbf{X}_j|}
\end{equation}
\end{definition}

\begin{definition}[\textbf{Graph Smoothing Level}]
\label{df.gsl}
The Graph Smoothing Level of the whole graph, $GSL$, is defined as: 
\begin{equation}
\small
GSL = \frac{1}{N}\sum_{i\in \mathcal{V}}NSL_i
\end{equation}
\end{definition}

$NSL_i$ measures the average similarities between node $i$ and every other node in the graph.
Corresponding to $NSL_i$, $GSL$ measures the average similarities between the node pairs in the graph.
Note that both metrics are positively correlated to the smoothness level.

\begin{figure*}[htp]
\centering  
\subfigure[The influence of $D_p$ to $GSL$.]{
\label{fig.smooth_nsl}
\includegraphics[width=0.3\textwidth]{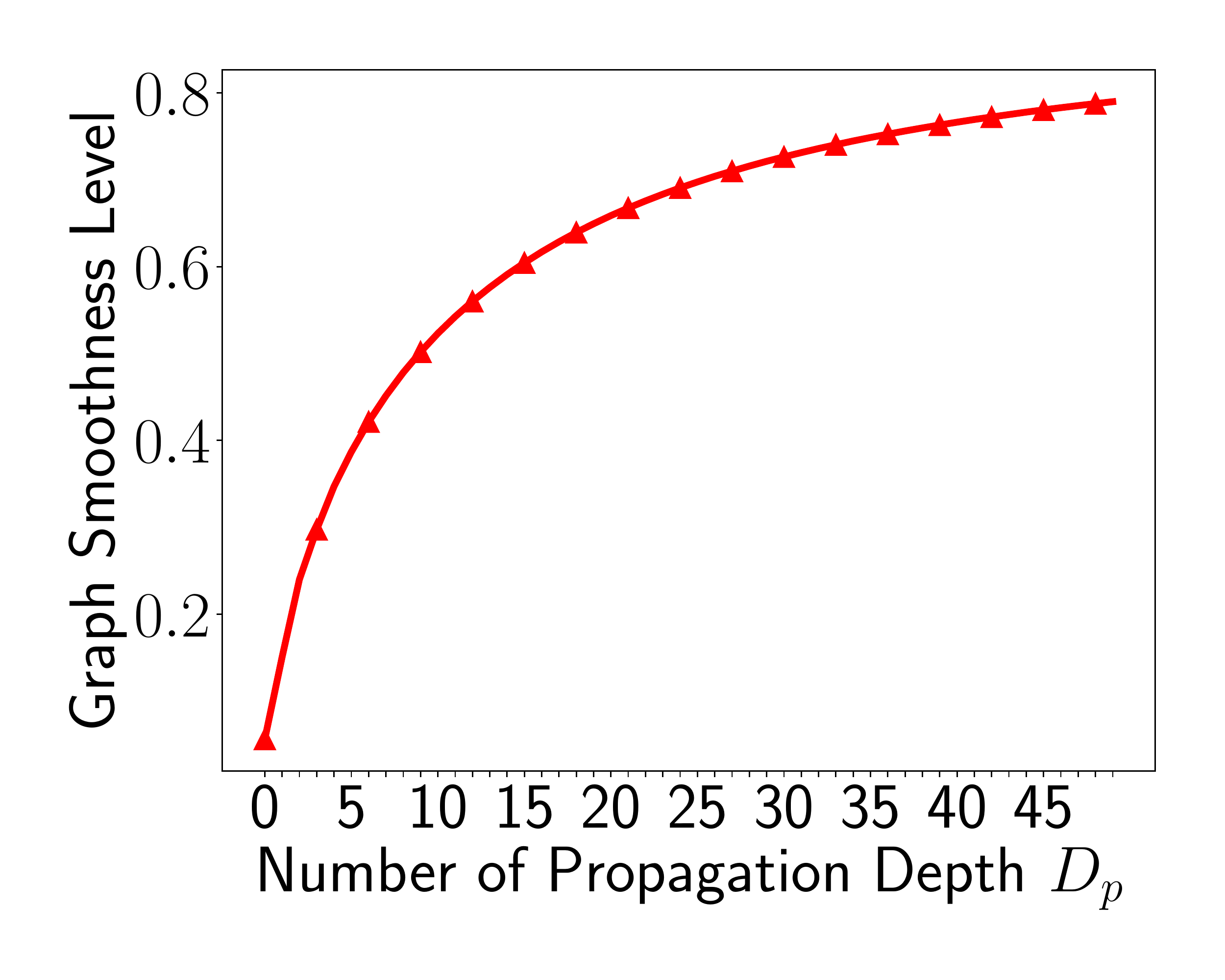}}
\subfigure[The influence of $D_p$ to model performance.]{
\label{fig.smooth_acc}
\includegraphics[width=0.3\textwidth]{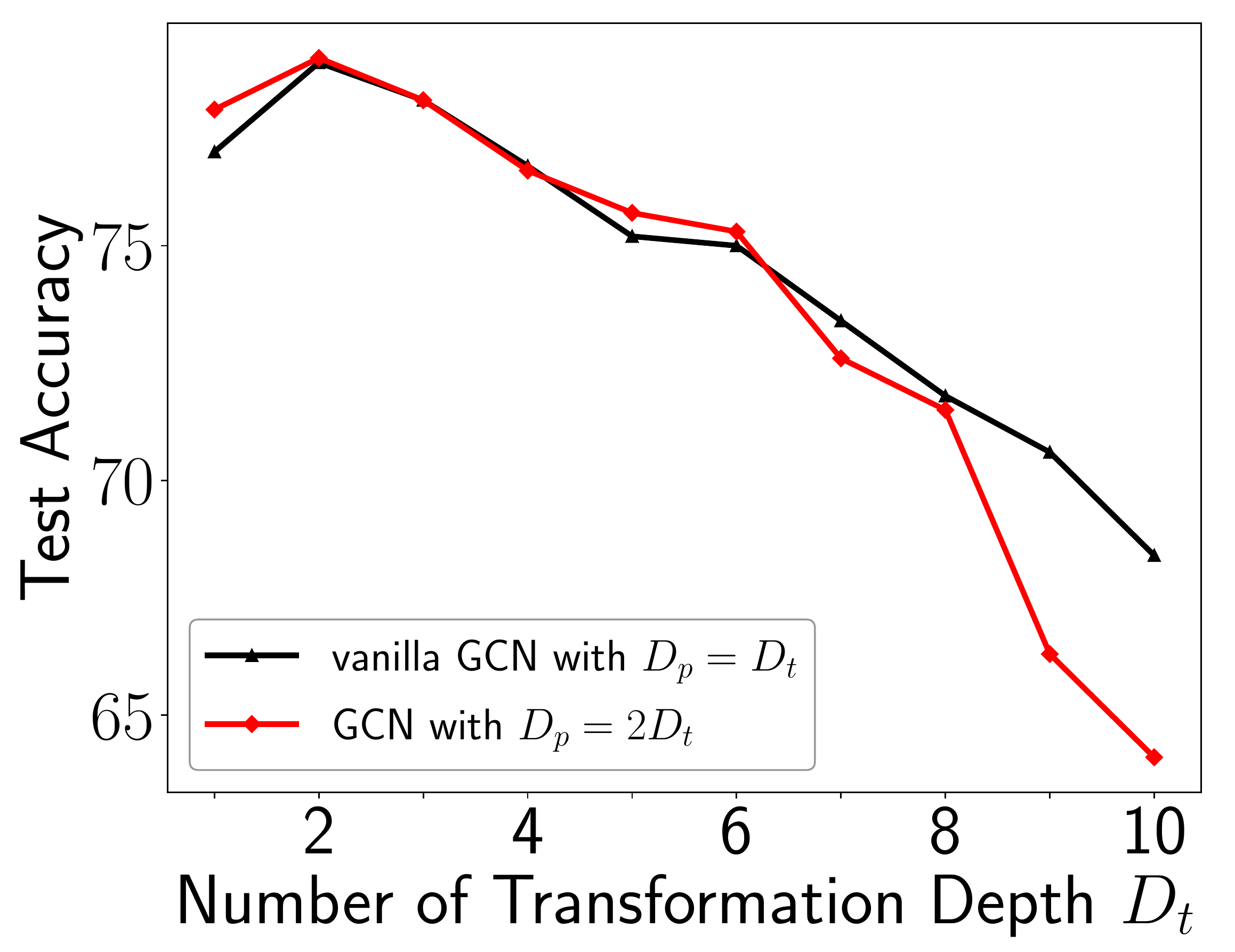}}
\subfigure[The influence of $D_t$ to model performance.]{
\label{fig.smooth_ak}
\includegraphics[width=0.3\textwidth]{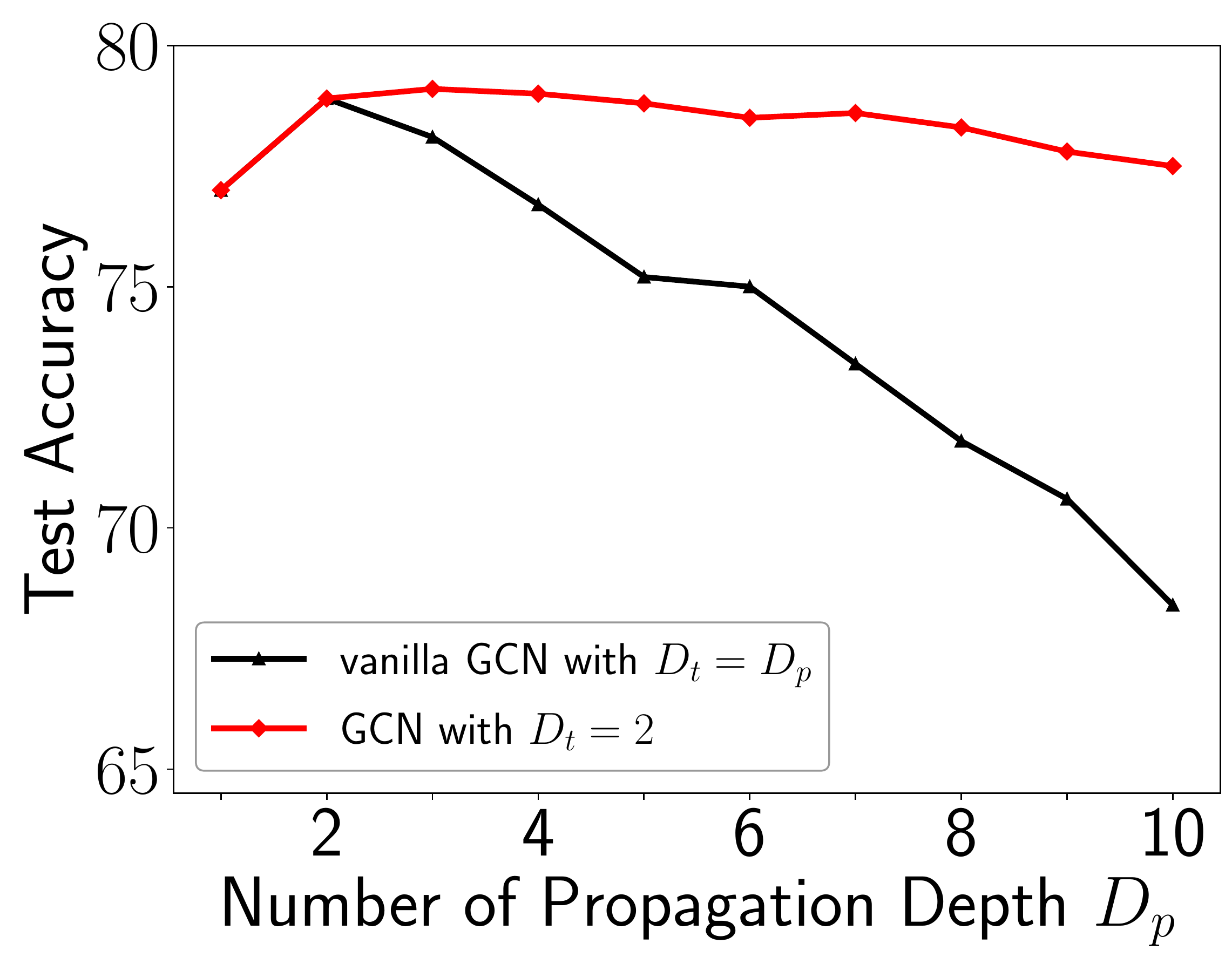}}
\vspace{-3mm}
\caption{\textit{Over-smoothing} is not the major contributor to the performance degradation of deep GNNs.}
\label{fig.oversmooth}
\vspace{-2mm}
\end{figure*}

\begin{figure}[htp]
	\centering
	\includegraphics[width=.65\linewidth]{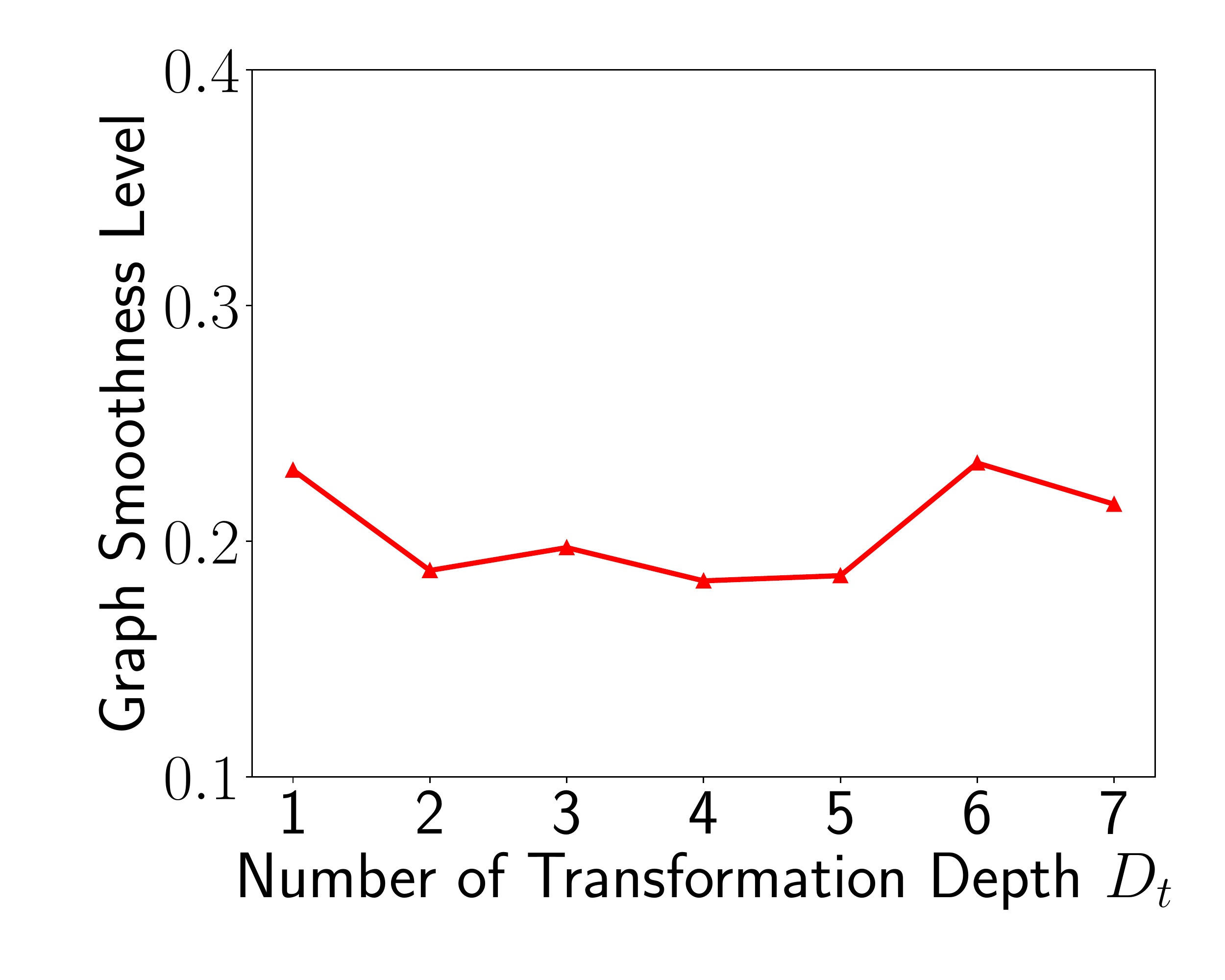}
	\vspace{-2mm}
	\caption{The influence of $D_t$ to $GSL$.}
	\vspace{-2mm}
	\label{fig.add_dt}
\end{figure}

\subsection{Review The \textit{Over-smoothing} Issue}
\label{sec.review_over_smoothing}
The \textit{over-smoothing} issue~\cite{li2018deeper} describes a phenomenon that the output representations of nodes become indistinguishable after applying the GNN model.
And the \textit{over-smoothing} issue always happens when a GNN model is stacked with many layers.

In the conventional \textbf{PTPT} architecture adopted by GCN, $D_p$ and $D_t$ are restrained to have the same value.
However, after the disentanglement of the graph convolution operation in this paper, it is feasible to analyze the respective effect caused by only enlarging $D_p$ or $D_t$.
In the analysis below, we will show that a huge $D_p$ is the actual reason behind the appearance of the \textit{over-smoothing} issue.

\subsubsection{Enlarging $D_p$}
In the \textbf{P} operation (Equation~\ref{eq_EP}), each time the normalized adjacency matrix $\mathbf{\hat{A}}$ multiplies with the input matrix $\mathbf{X}$, information one more hop away can be acquired for each node.
However, if we apply the \textbf{P} operations for infinite times, the node representations within the same connected component would reach a stationary state, leading to indistinguishable outputs.
Concretely, when adopting $\mathbf{\hat{A}}=\widetilde{\mathbf{D}}^{r-1}\tilde{\mathbf{A}}\widetilde{\mathbf{D}}^{-r}$, $\mathbf{\hat{A}}^{\infty}$ follows
\begin{equation}
\small
\label{eq.stationary}
\hat{\mathbf{A}}^{\infty}_{i,j} = \frac{(d_i+1)^r(d_j+1)^{1-r}}{2m+n},
\end{equation}
which shows that when $D_p$ approaches $\infty$, the influence from node $j$ to node $i$ is only determined by their node degrees. 
Correspondingly, the unique information of each node is fully smoothed, leading to indistinguishable representations, i.e., the \textit{over-smoothing} issue.

\subsubsection{Enlarging $D_t$}
Enlarging $D_t$, the number of the non-linear transformations, has no direct effect on the appearance of the \textit{over-smoothing} issue.
To support this claim, we evaluate the classification accuracies of a fully-disentangled \textbf{PPTT} GNN model, SGC, and the corresponding $GSL$ on the popular Cora~\cite{yang2016revisiting} dataset.
We enlarge SGC's $D_t$ while fix its $D_p=3$ to rule out the effects $D_p$ poses on the outputs.
The experimental results in Figure~\ref{fig.add_dt} show that the $GSL$ only fluctuates within a small interval.
There is no sign that the \textit{over-smoothing} issue would happen when only $D_t$ increases.

Carried out under the \textbf{PTPT} architecture,~\cite{Oono2020Graph} proves that the singular values of the learnable transformation matrix $\mathbf{W}$ and the non-linear activation function $\delta$ also correlate with the appearance of the \textit{over-smoothing} issue.
However, the assumptions~\cite{Oono2020Graph} adopts are rather rare in real-world scenarios (e.g., assuming a dense graph).
On the real-world dataset Cora, the above experiment shows that enlarging $D_t$ has no correlation with the \textit{over-smoothing} issue under the fully-disentangled \textbf{PPTT} architecture.

\subsection{Is \textit{Over-smoothing} the Major Cause?}
\label{sec.misconception}
Most previous studies~\cite{li2018deeper, zhang2019attributed} claim that the \textit{over-smoothing} issue is major cause for the failure of deep GNNs.
There have been lines of works aiming at designing deep GNNs.
For example, DropEdge~\cite{rong2019dropedge} randomly removes edges during training, and Grand~\cite{feng2020graph} randomly drops raw features of nodes before propagation.
Despite their ability to go deeper while maintaining or even achieving better predictive accuracy, the explanations for their effectiveness are misleading to some extent.
In this subsection, we empirically analyze whether the \textit{over-smoothing} issue is the major cause for the performance degradation of deep GNNs.

\subsubsection{Relations between $D_p$ and $GSL$} 
In Section~\ref{sec.review_over_smoothing}, we have illustrated that the \textit{over-smoothing} issue would always happen when $D_p$ approaches $\infty$.
However, the variation trend of the risk for the appearance of the \textit{over-smoothing} issue (i.e., $GSL$ of the graph) when $D_p$ grows from a relatively small value is not revealed.
To evaluate the single effect enlarging $D_p$ poses on the $GSL$ of the graph, we enlarge $D_p$ in a \textbf{PPTT} GNN model, SGC~\cite{wu2019simplifying}, and measure the $GSL$ of the intermediate node representations after all the \textbf{P} operations.
The experiment results on the Cora dataset are shown in Figure~\ref{fig.smooth_nsl}, and $GSL$ shows a monotonously increasing trend as $D_p$ grows.

\textit{Remark 1: The risk for the appearance of the \textit{over-smoothing} issue increases as $D_p$ grows.}

\subsubsection{Large $D_p$ Might Not Be The Major Cause} 
To investigate the relations between the smoothness level and the node classification accuracy, we increase the number of graph convolution operation in vanilla GCN ($D_p = D_t$) and a modified GCN with $\hat{\mathbf{A}}^{2}$ being the normalized adjacency matrix (i.e., $D_p = 2D_t$) on the PubMed dataset~\cite{yang2016revisiting}.
Supposing that the \textit{over-smoothing} issue is the major cause for the performance degradation of deep GNNs, the classification accuracy of the GCN with $D_p=2D_t$ should be much lower than the one of vanilla GCN.
The experimental results are shown in Figure~\ref{fig.smooth_acc}.
We can see that even with larger $D_p$ (i.e., higher smoothness level), GCN with $D_p = 2D_t$ always has similar classification accuracy with vanilla GCN ($D_p = D_t$) when $D_t$ ranges from $1$ to $8$, and the excessive number of \textbf{P} operations seems to begin dominating the performance decline only when $D_p$ exceeds $16$ ($2 \times 8$).
However, the performance of vanilla GCN starts to drop sharply when $D_p$ exceeds $2$, which is much smaller than $16$ (appearance of the performance gap in Figure~\ref{fig.smooth_acc}).

\textit{Remark 2: Considering the performance degradation of deep GNNs often happens even when the layer number is less than $10$, the over-smoothing issue might not be the major cause for it.}

\subsubsection{Large $D_t$ Dominates Performance Degradation} 
To dig out the actual limitation of deep GCNs, we adopt a two-layer GCN.
The normalized adjacency matrix $\hat{\mathbf{A}} $ is set to $(\widetilde{\mathbf{D}}^{-\frac{1}{2}}\widetilde{\mathbf{A}}\widetilde{\mathbf{D}}^{-\frac{1}{2}})^{\lfloor D_p/2 \rfloor}$ in the first layer of this GCN model; and $\hat{\mathbf{A}}$ is set to $(\widetilde{\mathbf{D}}^{-\frac{1}{2}}\widetilde{\mathbf{A}}\widetilde{\mathbf{D}}^{-\frac{1}{2}})^{\lceil D_p/2 \rceil}$ in the second layer.
This modified version of GCN will be referred to as ``GCN with $D_t=2$'' in the rest of the analysis.
We report the classification accuracies of vanilla GCN and ``GCN with $D_t=2$'' as $D_p$ increases in Figure~\ref{fig.smooth_ak}.
The experimental results show that the accuracy of ``GCN with $D_t = 2$'' does drop as $D_p$ grows, yet the decline is relatively small, while the accuracy of vanilla GCN (fix $D_p=D_t$) faces a sharp decline.
Thus, it can be inferred that although individually enlarging $D_p$ will increase the risk for the appearance of the \textit{over-smoothing} issue as the previous analysis shows, the performance is only slightly influenced. 
However, the performance will drop drastically if we simultaneously increase $D_t$.

\textit{Remark 3: Large $D_p$ will harm the classification accuracy of deep GNNs, yet the decline is relatively small.
On the contrary, large $D_t$ is the major cause for the performance degradation of deep GNNs.}

\begin{figure}[pbt!]
\centering  
\subfigure[Adding skip connections to MLP.]{
\label{fig.mlp}
\includegraphics[width=0.23\textwidth]{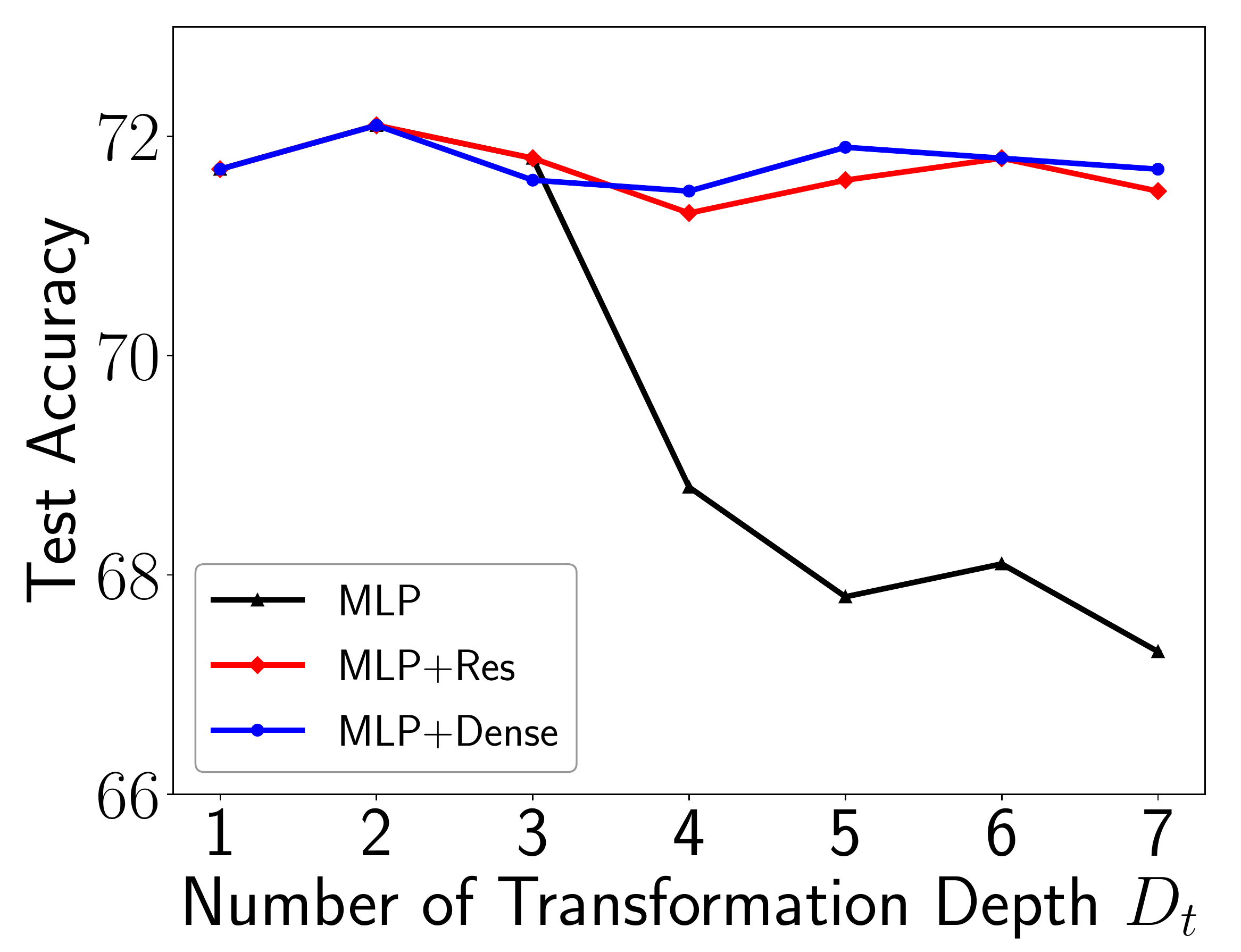}}
\subfigure[Adding skip connections to GCN.]{
\label{fig.res_acc}
\includegraphics[width=0.23\textwidth]{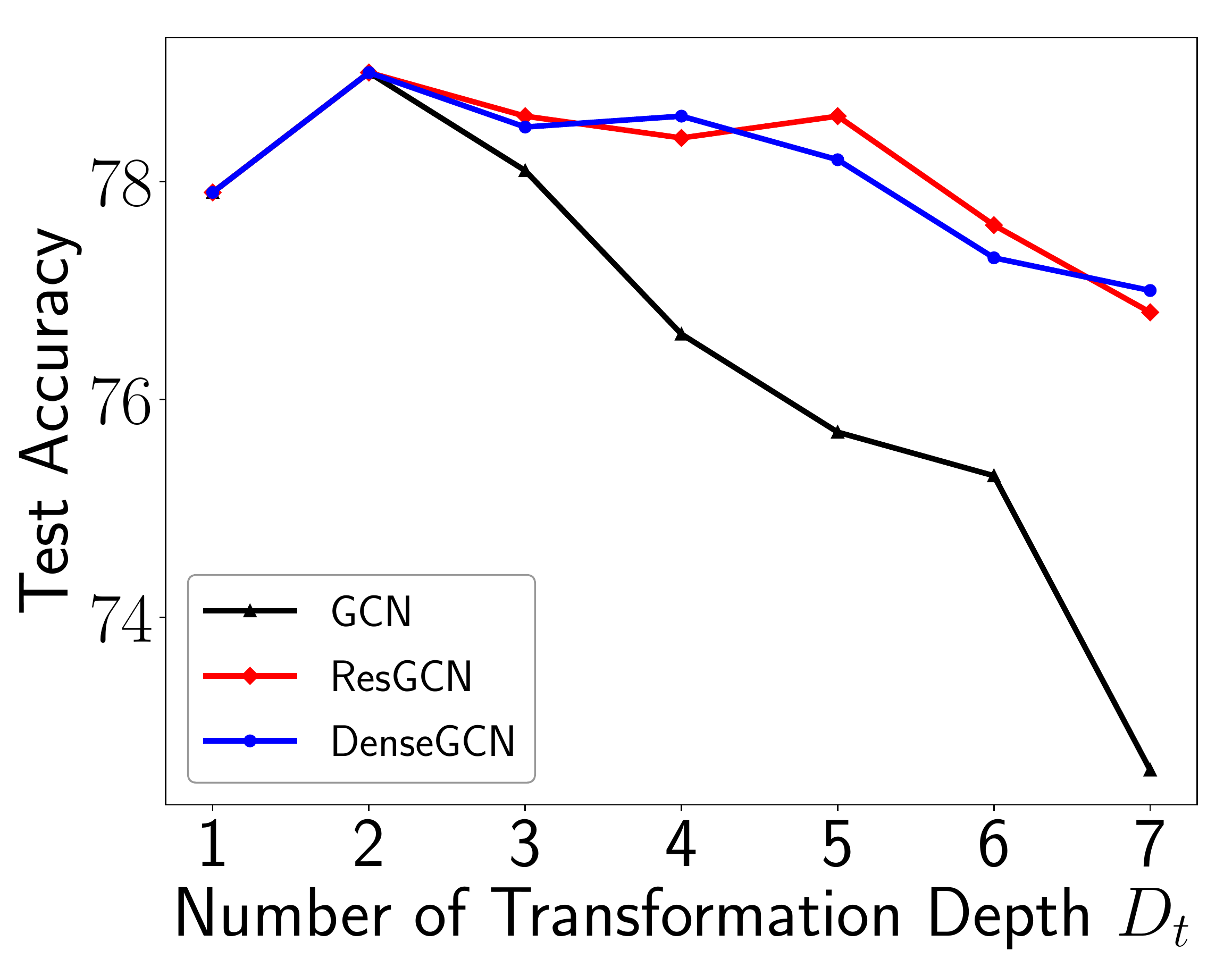}}
\vspace{-2mm}
\caption{The influence of skip connections in MLP and GCN.}
\label{fig.res_dense}
\vspace{-2mm}
\end{figure}

\section{What Is Behind Large $D_t$?}
\label{finding2}
To learn the fundamental limitation caused by large $D_t$, we first evaluate the classification accuracy of deep MLPs on the PubMed dataset and then extend the conclusions to deep GNNs.

\subsection{Deep MLPs Also Perform Bad} 
We evaluate the predictive accuracy of MLP as $D_t$, i.e., the number of MLP layers, grows on the PubMed dataset, and the black line in Figure~\ref{fig.mlp} shows the evaluation results.
It can be drawn from the results that the classification accuracy of MLP also decreases sharply when $D_t$ increases.
Thus, the performance degradation caused by large $D_t$ also exists in MLP.
It reminds us that the approaches to easing the training of deep MLPs might also help alleviate the performance degradation caused by large $D_t$ in deep GNNs.

\begin{figure}[tp!]
	\centering
	\includegraphics[width=.65\linewidth]{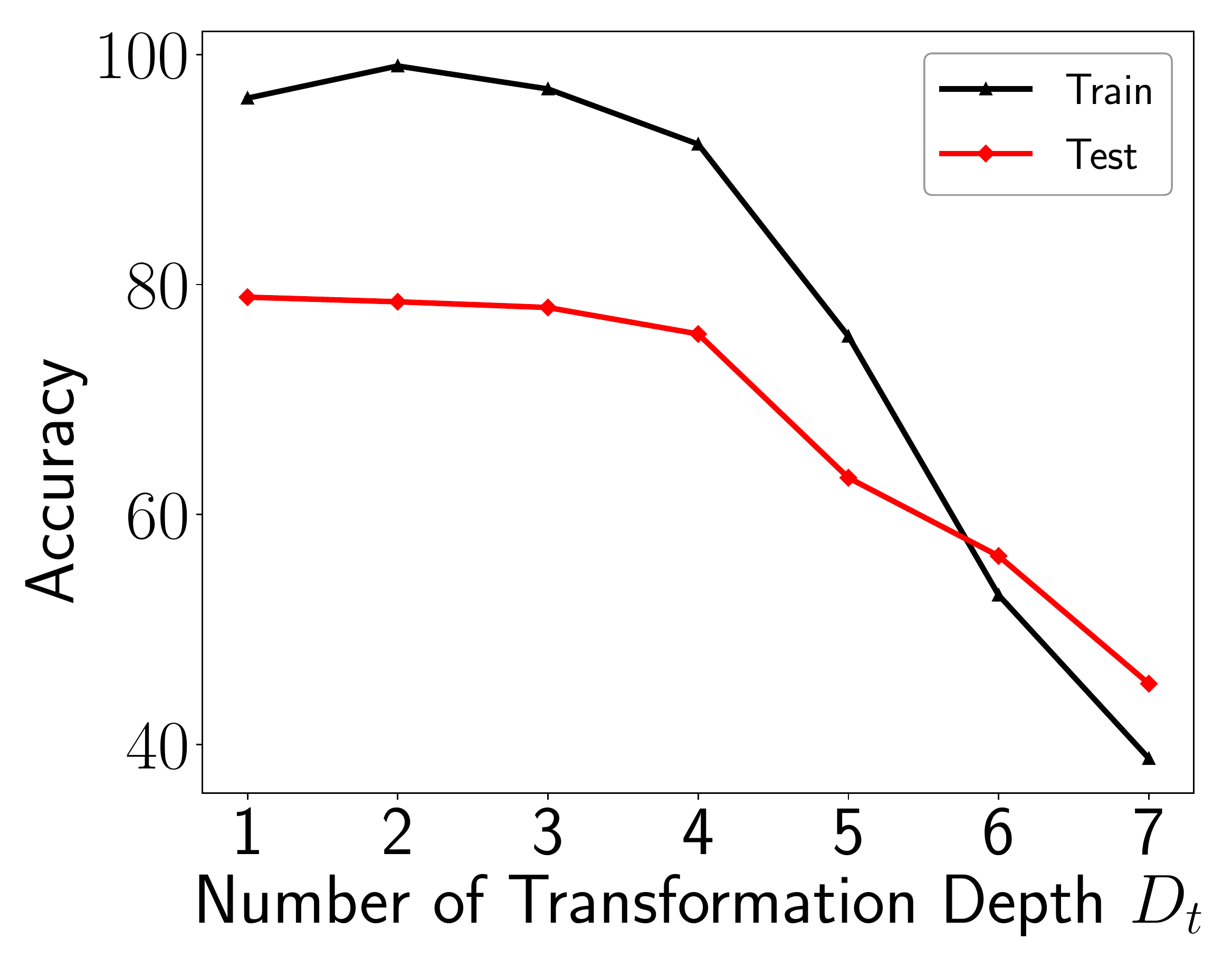}
	\vspace{-2mm}
	\caption{The influence of transformation depth in GCN.}
	\label{fig.overfit}
	\vspace{-2mm}
\end{figure}

\begin{figure*}[pbt!]
\centering  
\subfigure[Under \textbf{PPTT} architecture.]{
\label{fig.pptt_air}
\includegraphics[width=0.29\textwidth]{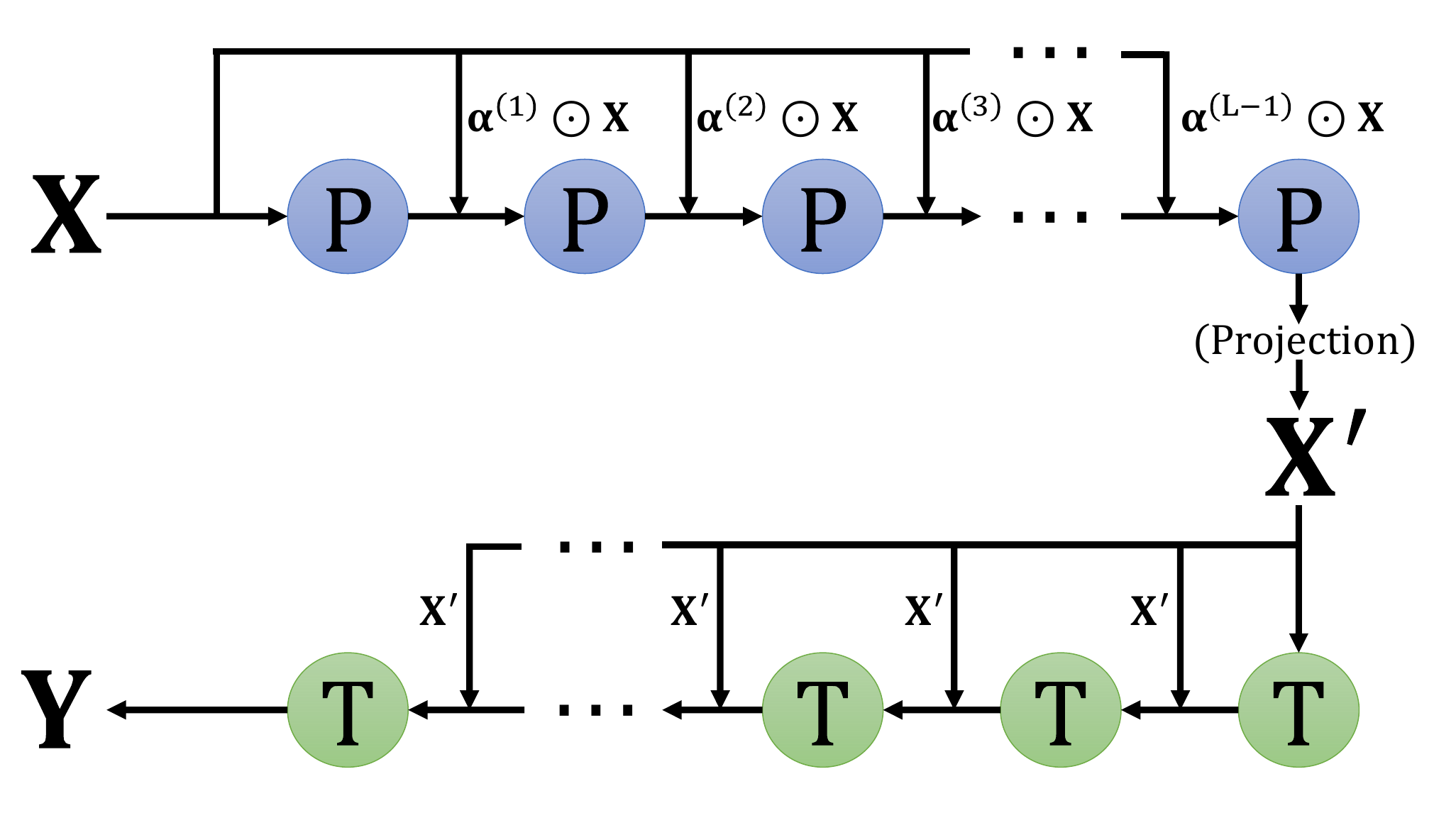}}
\hspace{1mm}
\subfigure[Under \textbf{TTPP} architecture.]{
\label{fig.ttpp_air}
\includegraphics[width=0.28\textwidth]{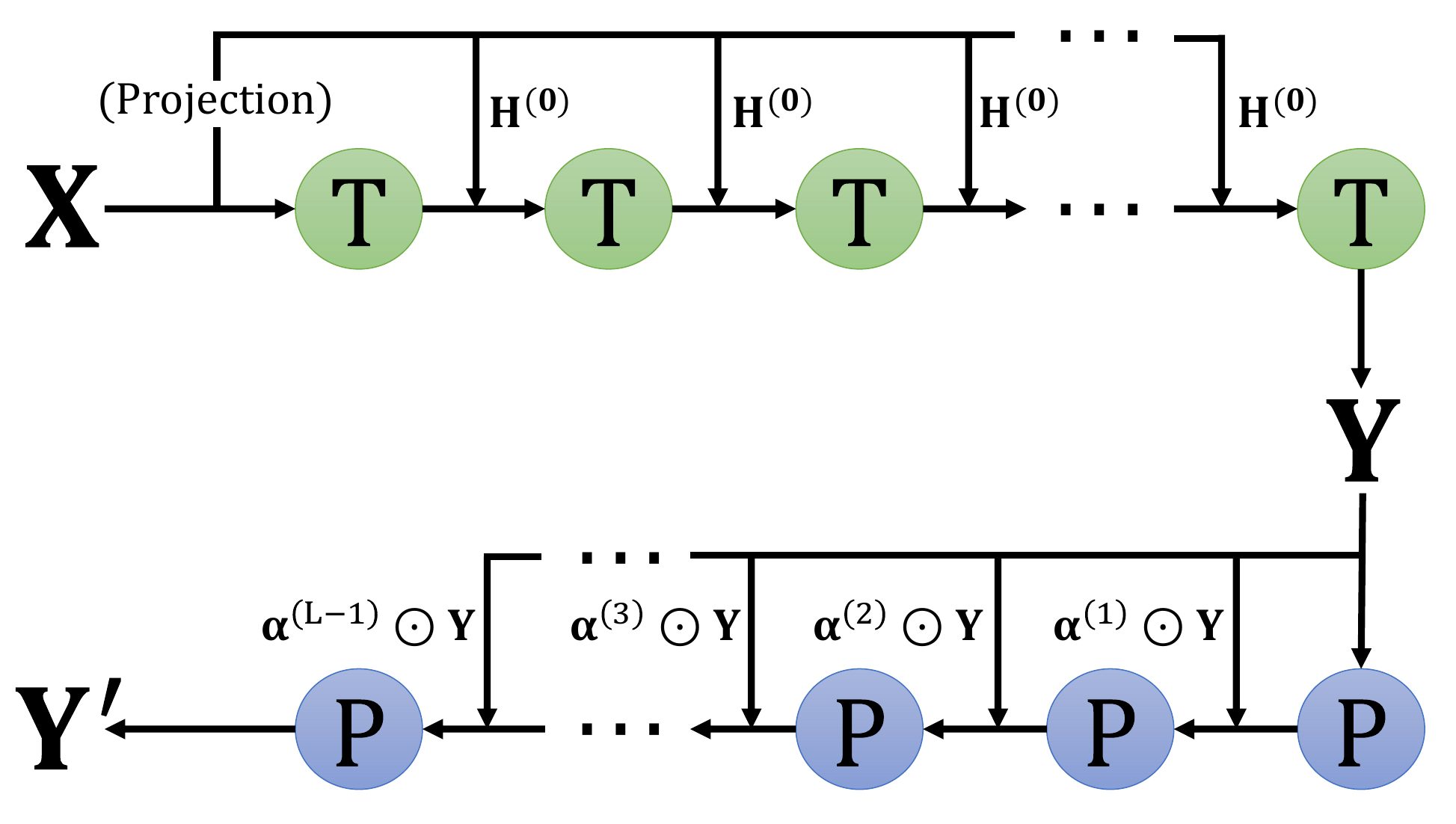}}
\hspace{1mm}
\subfigure[Under \textbf{PTPT} architecture.]{
\label{fig.ptpt_air}
\includegraphics[width=0.32\textwidth]{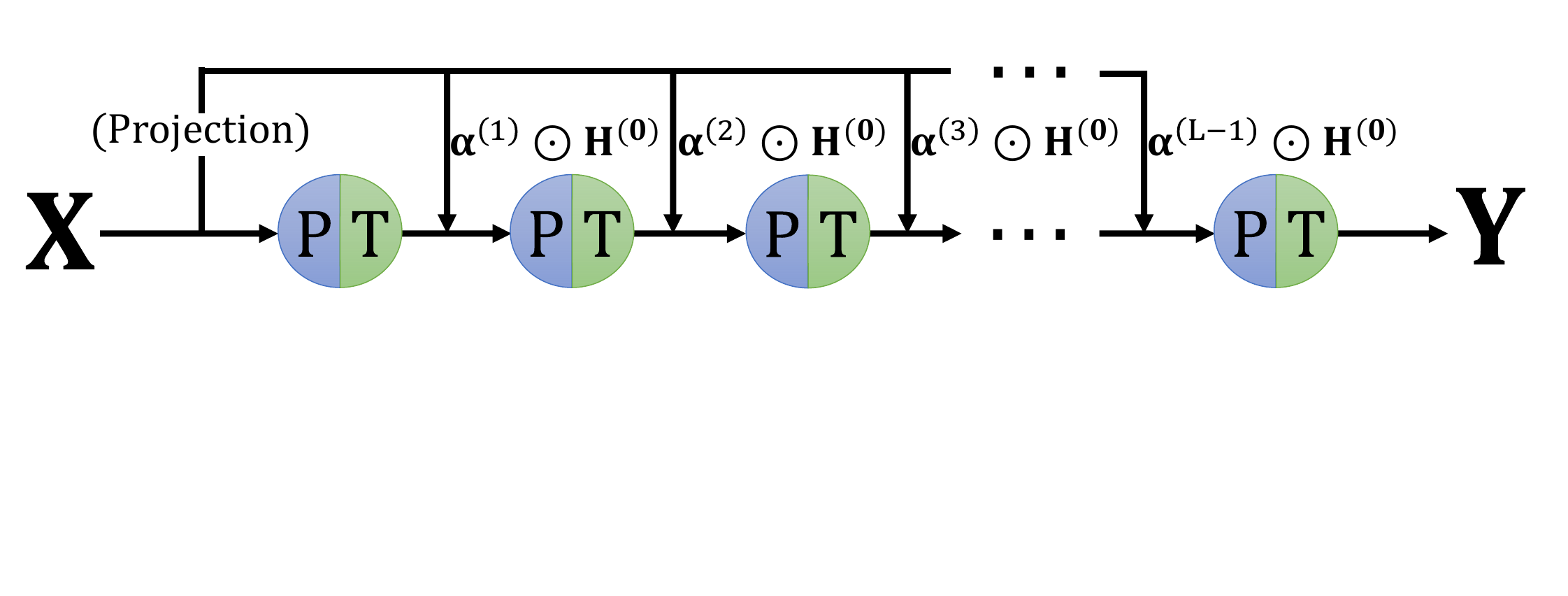}}
\vspace{-2mm}
\caption{Adopting AIR under different GNN architectures.}
\vspace{-2mm}
\end{figure*}

\subsection{Skip Connections Can Help} 
The widely-used approach that eases the training of deep MLPs is to add skip connections between layers~\cite{he2016deep, huang2017densely}.
Here, we add residual and dense connections to MLP and generate two MLP variants: ``MLP+Res'' and ``MLP+Dense'', respectively.
The classification accuracies of these two models as $D_t$ grows is shown in Figure~\ref{fig.mlp}.
Compared with plain deep MLP, the classification accuracies of both ``MLP+Res'' and ``MLP+Dense'' do not encounter massive decline when $D_t$ increases.
The evaluation results illustrate that adding residual or dense connections can effectively alleviate the performance degradation issue caused by large $D_t$ in deep MLPs.

\subsection{Extension to deep GNNs}
The empirical analysis in Section~\ref{sec.misconception} shows that large $D_t$ is the major cause for the performance degradation of deep GNNs.
However, it remains a pending question whether the widely-used approach to ease the training of deep MLPs can also alleviate the issue of deep GNNs. 
Thus, on the PubMed dataset, we evaluate the classification accuracies of ``ResGCN'' and ``DenseGCN'' in~\cite{li2019deepgcns}, which adds residual and dense connections between GCN layers, respectively.
The experimental results in Figure~\ref{fig.res_acc} illustrate that the performance decline of both ``ResGCN'' and ``DenseGCN'' can be nearly ignored compared to the massive performance decline of GCN.

\subsection{What do Skip Connections Help With Here?}
\subsubsection{Overfitting?}
It is pretty natural to guess that the \textit{overfitting} issue is the major cause for the performance degradation resulting from large $D_t$.
Concretely, the \textit{overfitting} issue comes from the case when an over-parametric model tries to fit a distribution with limited and biased training data, which results in a large generalization error.
To validate whether the \textit{overfitting} issue is behind large $D_t$, we evaluate the training accuracy and the test accuracy of GCN as the number of layers increases on the PubMed dataset.
The evaluation results are shown in Figure~\ref{fig.overfit}.
Figure~\ref{fig.overfit} shows that not only the test accuracy but also the training accuracy drops rapidly as the model layer exceeds $4$.
The decline of the training accuracy illustrates that the actual reason behind the performance degradation caused by large $D_t$ is not the \textit{overfitting} issue.

\subsubsection{Model Degradation!} 
The \textit{mode degradation} issue is first formally introduced in~\cite{he2016deep}.
It refers to the phenomenon that the model performance gets saturated and then degrades rapidly as the model grows deep.
What differentiates it from the \textit{overfitting} issue is that both the training accuracy and the test accuracy rather than just the test accuracy drops massively.
~\cite{he2016deep} does not explain the reasons for the appearance of the \textit{model degradation} issue but only presents a neat solution -- skip connections.
Many recent studies are trying to explore what is the leading cause for the \textit{model degradation} issue, and most of them probe into this problem from the gradient view~\cite{balduzzi2017shattered}.
The trend of the training accuracy and the test accuracy of deep GCN is precisely consistent with the phenomenon the \textit{model degradation} issue refers to.

\textit{Remark 4: The \textit{model degradation} issue behind large $D_t$ is the major cause for the performance degradation of deep GNNs.
Moreover, adding skip connections between layers can effectively alleviate the performance degradation of deep GNNs.}


\section{Adaptive Initial Residual (AIR)}
Under the above findings, we propose a plug-and-play module termed \textit{Adaptive Initial Residual} (AIR), which adds adaptive initial residual connections between the \textbf{P} and \textbf{T} operations.
The adaptive initial residual connections between the \textbf{P} operations aim to alleviate the \textit{over-smoothing} issue and take advantage of deeper information in a node-adaptive manner, while the main aim of the adaptive skip connections between the \textbf{T} operations is to alleviate the \textit{model degradation} issue.
We will introduce AIR in detail in the remainder of this section.
After that, the applications of AIR to three kinds of GNN architectures will also be discussed.

\subsection{AIR Between \textbf{P} and \textbf{T} Operations}
After the disentanglement of the graph convolution operation, nearly all the GNNs can be split into consecutive parts where each part is a sequence of continuous \textbf{P} operations or \textbf{T} operations.
Unlike skip connections in~\cite{he2016deep}, we directly construt a connection from the original inputs following~\cite{chen2020simple}.

\subsubsection{Between \textbf{P} Operations}
For a sequence of continuous \textbf{P} operations, we pull an adaptive fraction of input feature $\mathbf{H}^{(0)}_i=\mathbf{X}_i$ of node $i$ at each \textbf{P} operation.
Denote the representation of node $i$ at the $(l-1)$th operation as $\mathbf{H}^{(l-1)}_i$.
Then, the adaptive fraction $\alpha^{(l)}_i \in \mathbb{R}$ of node $i$ at $l$-th operation is computed as follows:
\begin{equation}
\small
    \alpha^{(l)}_i = \text{sigmoid}\left[\left(\mathbf{H}^{(l-1)}_i\big|\mathbf{H}^{(0)}_i\right)\mathbf{U}\right], 
\end{equation}
\noindent where $\mathbf{U}$ is a learnable vector that transforms the concatenated vector into a scalar.

All the positions of the $i$th row vector, $\bm{\alpha}^{(l)}_i$, of the adaptive weighting matrix $\bm{\alpha}^{(l)}$ are occupied by the same element $\alpha^{(l)}_i$:
\begin{equation}
\small
    \bm{\alpha}^{(l)}_i = [\alpha^{(l)}_i, \alpha^{(l)}_i, ..., \alpha^{(l)}_i]. \nonumber
\end{equation}

Then, for each $l \geq 2$, the $l$-th \textbf{P} operation equipped with AIR within a part can be formulated as follows:
\begin{equation}
\small
\label{eq.p_air}
    \mathbf{H}^{(l)} = \mathbf{\hat{A}}\left[(\mathbf{1}-\bm{\alpha}^{(l-1)}) \odot \mathbf{H}^{(l-1)}+\bm{\alpha}^{(l-1)} \odot \mathbf{H}^{(0)}\right],
\end{equation}
\noindent where $\mathbf{H}^{(l)}$ and $\mathbf{H}^{(l-1)}$ are the representation matrices after the $l$-th and the $(l-1)$th operation, respectively.
$\mathbf{1}$ is an all one matrix, and $\odot$ denotes the Hadamard product.
$\mathbf{\hat{A}}$ is the normalized adjacency matrix in Equation~\ref{eq_EP} and $\mathbf{H}^{(0)}$ is the input matrix of this part.

Equipped with AIR, the continuous \textbf{P} operations are equivalent to first computing the propagated inputs, $\left[\mathbf{H}^{0}, \mathbf{\hat{A}}\mathbf{H}^{(0)}, \mathbf{\hat{A}}^{2}\mathbf{H}^{(0)}, ...\right]$, then assigning each of them with learnable coefficient in a node-adaptive manner.
Under this view, it is evident that AIR can help alleviate the \textit{over-smoothing} issue.
The model can assign larger coefficients to deeper propagated inputs for nodes that require deep information.
For nodes that require only local information, the coefficients for deep propagated features can be assigned with values around zero.

\begin{table*}[tpb!]
\vspace{-1mm}
\caption{Test accuracy on the node classification task. ``OOM'' means ``out of memory''.}
\vspace{-1mm}
\centering
{
\noindent
\renewcommand{\multirowsetup}{\centering}
\resizebox{0.75\linewidth}{!}{
\begin{tabular}{ccccccc}
\toprule
\textbf{Methods} & \textbf{Cora} & \textbf{Citeseer} & \textbf{PubMed} & \textbf{ogbn-arxiv} & \textbf{ogbn-products} & \textbf{ogbn-papers100M}\\
\midrule
GCN& 81.8$\pm$0.5 & 70.8$\pm$0.5 &79.3$\pm$0.7 & 71.74$\pm$0.29 & OOM & OOM \\
GraphSAGE& 79.2$\pm$0.6 & 71.6$\pm$0.5 & 77.4$\pm$0.5 &71.49$\pm$0.27 & \underline{78.29$\pm$0.16} & 64.83$\pm$0.15 \\

JK-Net& 81.8$\pm$0.5  & 70.7$\pm$0.7 & 78.8$\pm$0.7 & 72.19$\pm$0.21 & OOM & OOM   \\
ResGCN& 81.2$\pm$0.5  & 70.8$\pm$0.4 & 78.6$\pm$0.6 & 72.62$\pm$0.37 & OOM & OOM    \\
\midrule
\textbf{GCN+AIR} & 83.2$\pm$0.7 & 71.6$\pm$0.6 & 80.2$\pm$0.7 & \textbf{72.69$\pm$0.28} & OOM & OOM \\
\midrule

APPNP& 83.3$\pm$0.5 & 71.8$\pm$0.5 & 80.1$\pm$0.2 & 71.83$\pm$0.31 & OOM & OOM \\
AP-GCN& 83.4$\pm$0.3& 71.3$\pm$0.5& 79.7$\pm$0.3 &  71.92$\pm$0.23 & OOM & OOM \\
DAGNN & \textbf{84.4$\pm$0.5} & \underline{73.3$\pm$0.6} & 80.5$\pm$0.5  & 72.09$\pm$0.25 & OOM & OOM \\
\midrule
\textbf{APPNP+AIR} & 83.8$\pm$0.6 & \textbf{73.4$\pm$0.5} & \underline{81.0$\pm$0.6}  & 72.16$\pm$0.22 & OOM & OOM \\
\midrule

SGC & 81.0$\pm$0.2 & 71.3$\pm$0.5 & 78.9$\pm$0.5  &71.42$\pm$0.26 & 75.94$\pm$0.22 & 63.29$\pm$0.19\\
SIGN & 82.1$\pm$0.3 & 72.4$\pm$0.8 & 79.5$\pm$0.5 & 71.95$\pm$0.11 & 76.83$\pm$0.39 & 64.28$\pm$0.14\\
S$^2$GC& 82.7$\pm$0.3 & 73.0$\pm$0.2 &79.9$\pm$0.3 & 71.83$\pm$0.31 & 77.13$\pm$0.24 & 64.73$\pm$0.21\\
GBP& 83.9$\pm$0.7 & 72.9$\pm$0.5 & 80.6$\pm$0.4 &72.24$\pm$0.23 & 77.68$\pm$0.25 & \underline{65.24$\pm$0.13}\\
\midrule 
\textbf{SGC+AIR} & \underline{84.0$\pm$0.6} & 72.0$\pm$0.5 & \textbf{81.1$\pm$0.6}  & \underline{72.67$\pm$0.28} & \textbf{81.44$\pm$0.16} & \textbf{67.23$\pm$0.2} \\
\bottomrule
\end{tabular}}}
\label{table.performance}
\end{table*}

\subsubsection{Between \textbf{T} Operations}
Different from the adaptive initial residual connections between \textbf{P} operations, the connections between \textbf{T} operations exclude the learnable coefficients for the input feature, and a fixed one is adopted instead since these two ways are almost equivalent in this scenario. 

Similar to the \textbf{P} operations introduced above, for each $l \geq 2$, the $l$-th \textbf{T} operation equipped with AIR within a part can be formulated as follows:
\begin{equation}
\small
    \mathbf{H}^{(l)} = \sigma\left[\left(\mathbf{H}^{(l-1)}+\mathbf{H}^{(0)}\right)\mathbf{W}\right],
\end{equation}
\noindent where $\mathbf{W}$ is the learnable transformation matrix and $\sigma$ is the non-linear activation function.
Note that the dimensions of the inputs and the latent representations might be different.
A linear projection layer is adopted to transform the inputs to the given dimension under such scenarios.

The initial residual connections~\cite{chen2020simple} that we adopt here share the same intuition with the residual connections in~\cite{he2016deep} that a deep model should at least achieve the same performance as a shallow one.
Thus, adopting the \textbf{T} operation with AIR is expected to alleviate the \textit{model degradation} issue caused by large $D_t$.

\subsection{Applications to Existing GNNs}
\subsubsection{\textbf{PPTT} and \textbf{TTPP}}
For the disentangled \textbf{PPTT} and \textbf{TTPP} GNN architectures, the model can be split into two parts, the one of which only consists of the \textbf{P} operations, the other one of which only consists of the \textbf{T} operations.
Thus, the \textbf{P} and \textbf{T} operations can be easily replaced by the \textbf{P} and \textbf{T} operations equipped with AIR introduced in the above subsection.

The \textbf{P} operation equipped with AIR pulls an adaptive amount of the original inputs directly over and fuses it with the representation matrix generated by the previous \textbf{P} operation. 
The fused matrix is considered as the new input to the $l$-th \textbf{P} operation.
The \textbf{T} operation equipped with AIR adds the outputs of the previous \textbf{T} operation and a fixed amount of the original inputs.

The examples of adopting AIR under the \textbf{PPTT} and \textbf{TTPP} architectures are given in Figure~\ref{fig.pptt_air} and~\ref{fig.ttpp_air}, respectively.

\subsubsection{\textbf{PTPT}}
The AIR for the \textbf{PTPT} GNN architecture is slightly different from the one for the \textbf{PPTT} and \textbf{TTPP} GNN architectures since the \textbf{P} and \textbf{T} operations are entangled in the \textbf{PTPT} architecture.
The GNN models under the \textbf{PTPT} architecture can be deemed as a sequence of the graph convolution operations.
Thus, we construct adaptive initial residual connections directly between the graph convolution operations.

If the dimensions of the inputs and latent representations are different, a linear projection layer is used to transform the inputs to the given dimension.
For each $l \geq 2$, the $l$-th graph convolution operation equipped with AIR can be formulated as:
\begin{equation}
\small
    \mathbf{H}^{(l)} = \sigma\left[\mathbf{\hat{A}}\left((\mathbf{1}-\bm{\alpha}^{(l-1)}) \odot \mathbf{H}^{(l-1)}+\bm{\alpha}^{(l-1)} \odot \mathbf{H}^{(0)}\right)\mathbf{W}\right],
\end{equation}
\noindent which is equivalent to applying the \textbf{P} operation equipped with AIR (Equation~\ref{eq.p_air}) and the \textbf{T} operation (Equation~\ref{eq_ET}) consecutively.

Figure~\ref{fig.ptpt_air} provides an overview of how to adopt AIR under the \textbf{PTPT} GNN architecture.

\section{AIR Evaluation}
In this section, we evaluate the proposed AIR under three different GNN architectures.
We adapt AIR to three GNN models: SGC~\cite{wu2019simplifying} (\textbf{PPTT}), APPNP~\cite{klicpera2018predict} (\textbf{TTPP}), and GCN~\cite{kipf2016semi} (\textbf{PTPT}), which are representative GNN models for the three GNN architectures, respectively.
Firstly, we introduce the utilized datasets and the experimental setup.
Then, we compare SGC+AIR, APPNP+AIR, and GCN+AIR with baseline methods regarding predictive accuracy, ability to go deep, robustness to graph sparsity, and efficiency.

\begin{figure*}[tbp!]
\centering  
\subfigure[fix $D_t$ change $D_p$ under \textbf{PPTT} and \textbf{TTPP}]{
\label{fig.dp_change}
\includegraphics[width=0.3\textwidth]{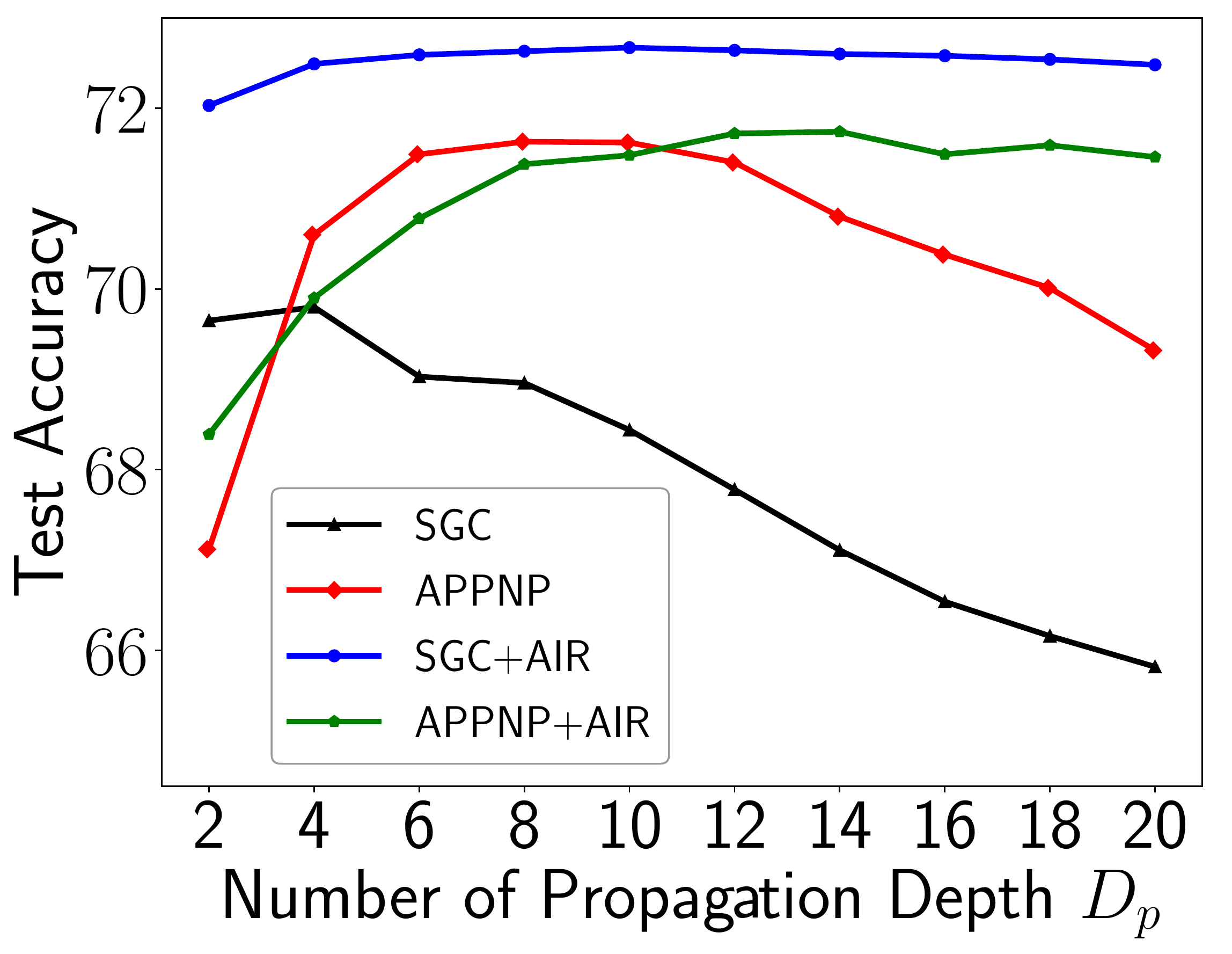}}
\subfigure[fix $D_p$ change $D_t$ under \textbf{PPTT} and \textbf{TTPP}]{
\label{fig.dt_change}
\includegraphics[width=0.3\textwidth]{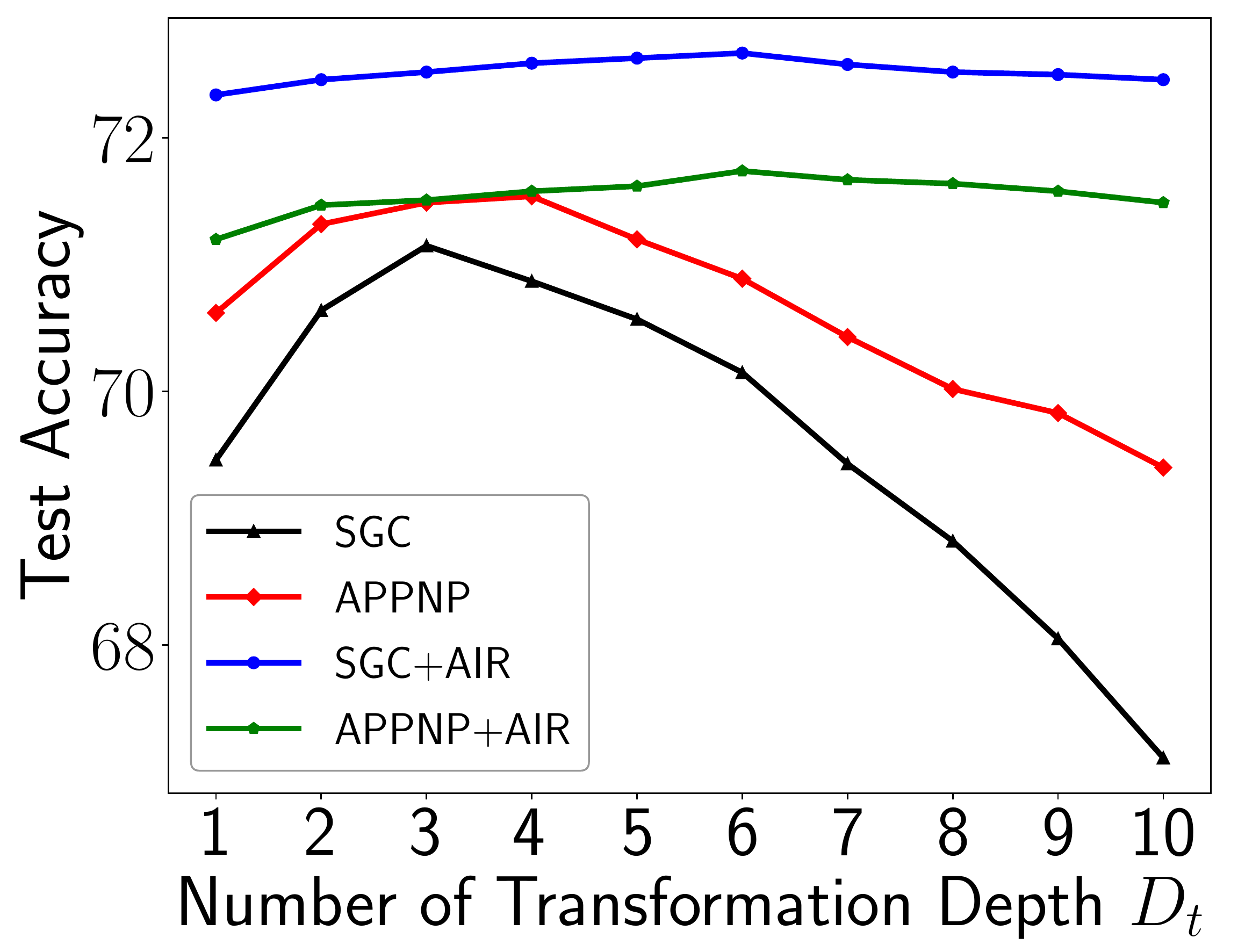}}
\subfigure[change $D_p$ ($=D_t$) under \textbf{PTPT}]{
\label{fig.dpdt_change}
\includegraphics[width=0.3\textwidth]{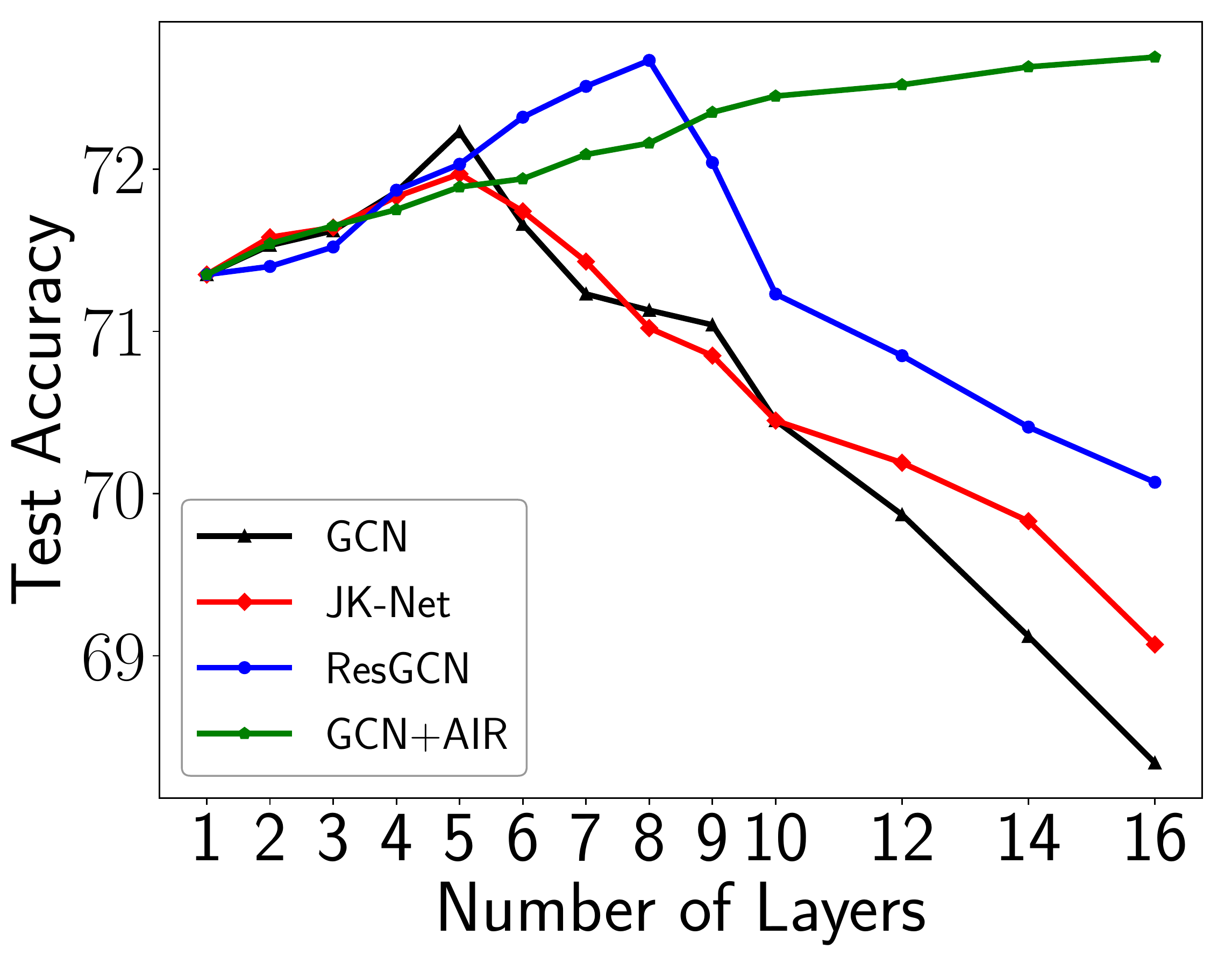}}
\vspace{-1mm}
\caption{Test accuracy with varied $D_p$s and $D_t$s under different GNN architectures.}
\label{fig.depth_ana}
\vspace{-1mm}
\end{figure*}

\subsection{Experimental Settings}
\label{sec:settings}
\noindent\textbf{Datasets.}
We adopt the three popular citation network datasets (Cora, Citeseer, PubMed)~\cite{DBLP:journals/aim/SenNBGGE08} and three large OGB datasets (ogbn-arxiv, ogbn-products, ogbn-papers100M)~\cite{hu2020ogb} to evaluate the predictive accuracy of each method on the node classification task.
Table~\ref{datasets} in Appendix~\ref{app:datasets} presents an overview of these six datasets.

\noindent\textbf{Baselines.}
We choose the following baselines: GCN~\cite{kipf2016semi}, GraphSAGE~\cite{hamilton2017inductive}, JK-Net~\cite{xu2018representation}, ResGCN~\cite{li2019deepgcns}, APPNP~\cite{klicpera2018predict}, AP-GCN~\cite{spinelli2020adaptive}, DAGNN~\cite{liu2020towards}, SGC~\cite{wu2019simplifying},      SIGN~\cite{frasca2020sign}, S$^2$GC~\cite{zhu2021simple}, and GBP~\cite{DBLP:conf/nips/ChenWDL00W20}.
The hyperparameter details for SGC+AIR, APPNP+AIR, GCN+AIR, and all the baseline methods can be found in Appendix~\ref{app_parameter}.

\subsection{End-to-End Comparison} 
The evaluation results of GCN+AIR, APPNP+AIR, SGC+AIR, and all the compared baselines on the six datasets are summarized in Table~\ref{table.performance}.
Equipped with AIR, GCN, APPNP, and SGC all achieve far better performance than their respective original version.
For example, the predictive accuracies of GCN+AIR, APPNP+AIR, and SGC+AIR exceed the one of their original version by $0.9\%$, $0.9\%$, and $2.2\%$ on the PubMed dataset, respectively.
Further, GCN+AIR, APPNP+AIR, and SGC+AIR also outperform or achieve comparable performance with state-of-the-art baseline methods within their own GNN architectures.

It is worth noting that the performance advantage of SGC+AIR over compared baseline methods on the two largest datasets, ogbn-products, and ogbn-papers100M, is more significant than the one on the smaller datasets.
This contrast is because AIR enables both larger $D_p$ and $D_t$ in GNNs, which helps the model exploit more valuable deep information on large datasets.

\subsection{Analysis of model depth}
In this subsection, we conduct experiments on the ogbn-arxiv dataset to validate that simple GNN methods can support large $D_p$ and large $D_t$ when equipped with AIR.

We first increase $D_p$ or $D_t$ individually under the \textbf{PPTT} and \textbf{TTPP} architectures.
In Figure~\ref{fig.dp_change}, we fix $D_t=3$ and increase $D_p$ from $1$ to $20$.
Figure~\ref{fig.dp_change} shows that SGC+AIR outperforms SGC throughout the experiment, and the performance of APPNP+AIR begins to exceed the one of APPNP when $D_p$ surpasses $10$.
The experimental results clearly illustrate that AIR can significantly reduce the risk for the appearance of the \textit{over-smoothing} issue.

In Figure~\ref{fig.dt_change}, we fix $D_p$ to $10$ and increase $D_t$ from $1$ to $10$.
While SGC and APPNP both encounter significant performance drop as $D_t$ exceeds $4$, the predictive accuracies of SGC+AIR and APPNP+AIR maintains or even becomes higher when $D_t$ grows.
This sharp contrast illustrates that AIR can greatly alleviate the \textit{model degradation} issue.
Thus, equipped with AIR, SGC and APPNP can better exploit the deep information and achieve higher predictive accuracy.

Under the \textbf{PTPT} architecture, we increase both $D_p$ and $D_t$ since the \textbf{P} and \textbf{T} operations are entangled in this architecture.
The experimental results are shown in Figure~\ref{fig.dpdt_change}.
Compared with baseline methods GCN, JK-Net, and ResGCN, GCN+AIR shows a steady increasing trend in predictive accuracy as the number of layers grows, which again validates the effectiveness of AIR.

\subsection{Performance-Efficiency Analysis}
\label{effi}

\begin{figure}
	\centering
	\includegraphics[width=.65\linewidth]{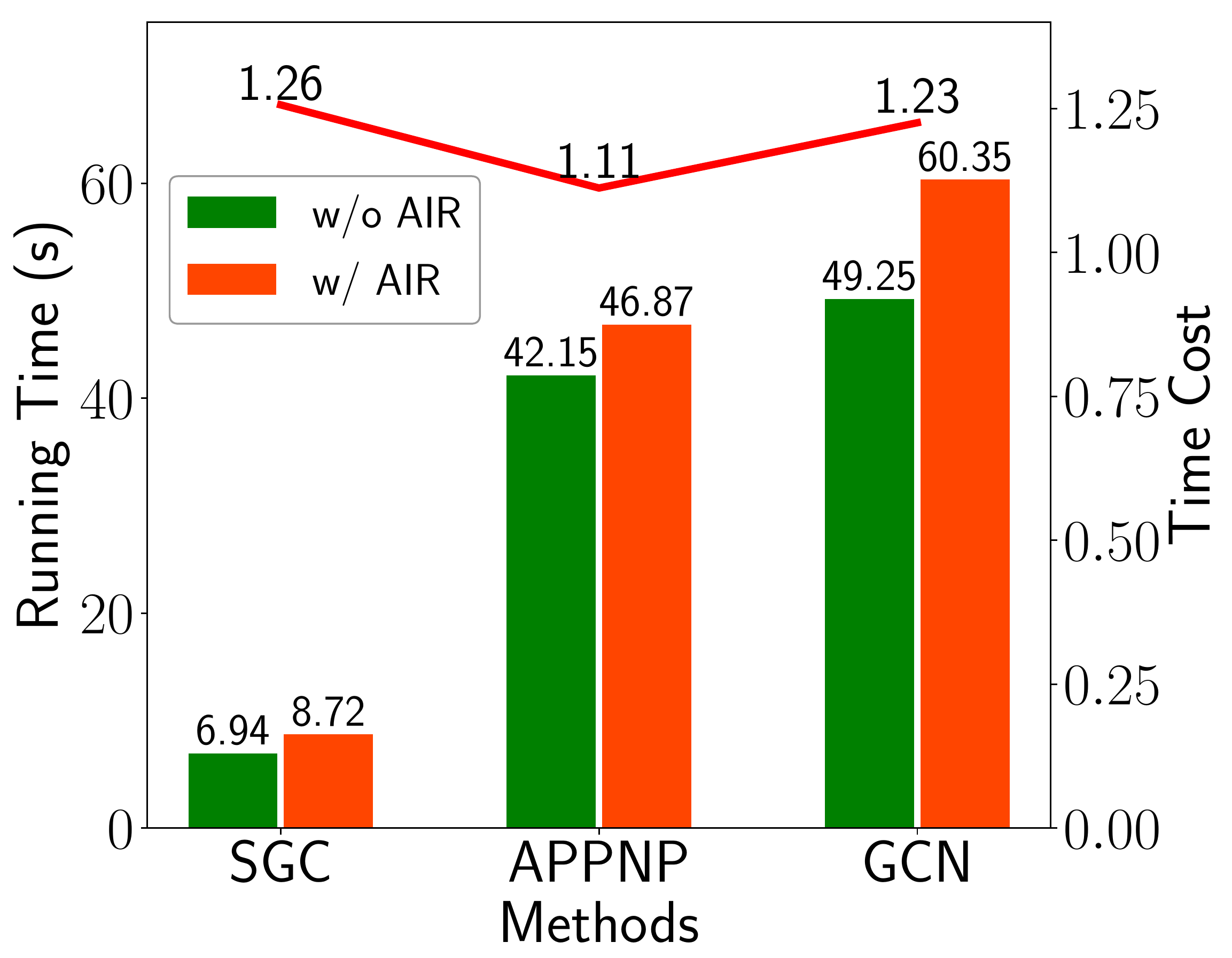}
  	\vspace{-1mm}
	\caption{Efficiency comparison on the ogbn-arxiv dataset.}
	\label{fig:efficiency}
 	\vspace{-2mm}
\end{figure}

In this subsection, we evaluate the efficiency of AIR on the ogbn-arxiv dataset.
Here we only report the training time since the training stage always consumes the most resources in real-world scenarios.
The training time results of SGC, APPNP, and GCN with or without AIR are shown in Figure~\ref{fig:efficiency}.
$D_p$ and $D_t$ are fixed to $3$ for all the compared methods, and each method is trained for $500$ epochs.

The experimental results in Figure~\ref{fig:efficiency} show that the additional time costs introduced by AIR vary from only $11\%$ to $26\%$, based on the training time of their respective original versions.
The time costs associated with AIR are perfectly accepted compared with the considerable performance improvement shown in Table~\ref{table.performance}.
Further, Figure~\ref{fig:efficiency} illustrates that SGC, which belongs to the \textbf{PPTT} architecture, consumes much less training time than APPNP and GCN, which belong to the \textbf{TTPP} and \textbf{PTPT} architecture, respectively.

\section{Conclusion}
In this paper, we perform an empirical analysis of current GNNs and find the root cause for the performance degradation of deep GNNs: the \textit{model degradation} issue introduced by large transformation depth ($D_t$). 
The \textit{over-smoothing} issue introduced by large propagation depth ($D_p$) does harm the predictive accuracy.
However, we find that the \textit{model degradation} issue always happens much earlier than the \textit{over-smoothing} issue when $D_p$ and $D_t$ increase at similar speeds.
Based on the above analysis, we present Adaptive Initial Residual (AIR), a plug-and-play module that helps GNNs simultaneously support large propagation and transformation depth.
Extensive experiments on six real-world graph datasets demonstrate that simple GNN methods equipped with AIR outperform state-of-the-art GNN methods, and the additional time costs associated with AIR can be ignored.

\begin{acks}
This work is supported by NSFC (No. 61832001, 61972004), Beijing Academy of Artificial Intelligence (BAAI), and PKU-Tencent Joint Research Lab. Wentao Zhang and Zeang Sheng contributed equally to this work, and Bin Cui is the corresponding author.
\end{acks}


\bibliographystyle{ACM-Reference-Format}
\bibliography{sample-base}

\clearpage

\appendix
\begin{table*}[tpb!]
\small
\centering
\caption{Overview of datasets.}
\vspace{-2mm}
\label{datasets}
\resizebox{.80\linewidth}{!}{
\begin{tabular}{cccccccc}
\toprule
\textbf{Dataset}&\textbf{\#Nodes}& \textbf{\#Features}&\textbf{\#Edges}&\textbf{\#Classes}&\textbf{\#Train/Val/Test}\\
\midrule
Cora& 2,708 & 1,433 &5,429&7& 140/500/1,000\\
Citeseer& 3,327 & 3,703&4,732&6& 120/500/1,000\\
Pubmed& 19,717 & 500 &44,338&3& 60/500/1,000\\
ogbn-arxiv& 169,343 & 128 & 1,166,243 & 40 &  91K/30K/47K 
\\
ogbn-products& 2,449,029 & 100 & 61,859,140 & 47 &  
196K/49K/2,204K\\
ogbn-papers100M & 111,059,956 & 128 & 1,615,685,872 & 172 & 
1,207K/125K/214K\\
\bottomrule
\end{tabular}}
\end{table*}

\begin{figure*}[tpb!]
\centering  
\subfigure[Fixing $D_t=D_p$ degrades the performance badly when $D_p$ becomes large on the Cora dataset.]{
\label{fig:coupled}
\includegraphics[width=0.3\textwidth]{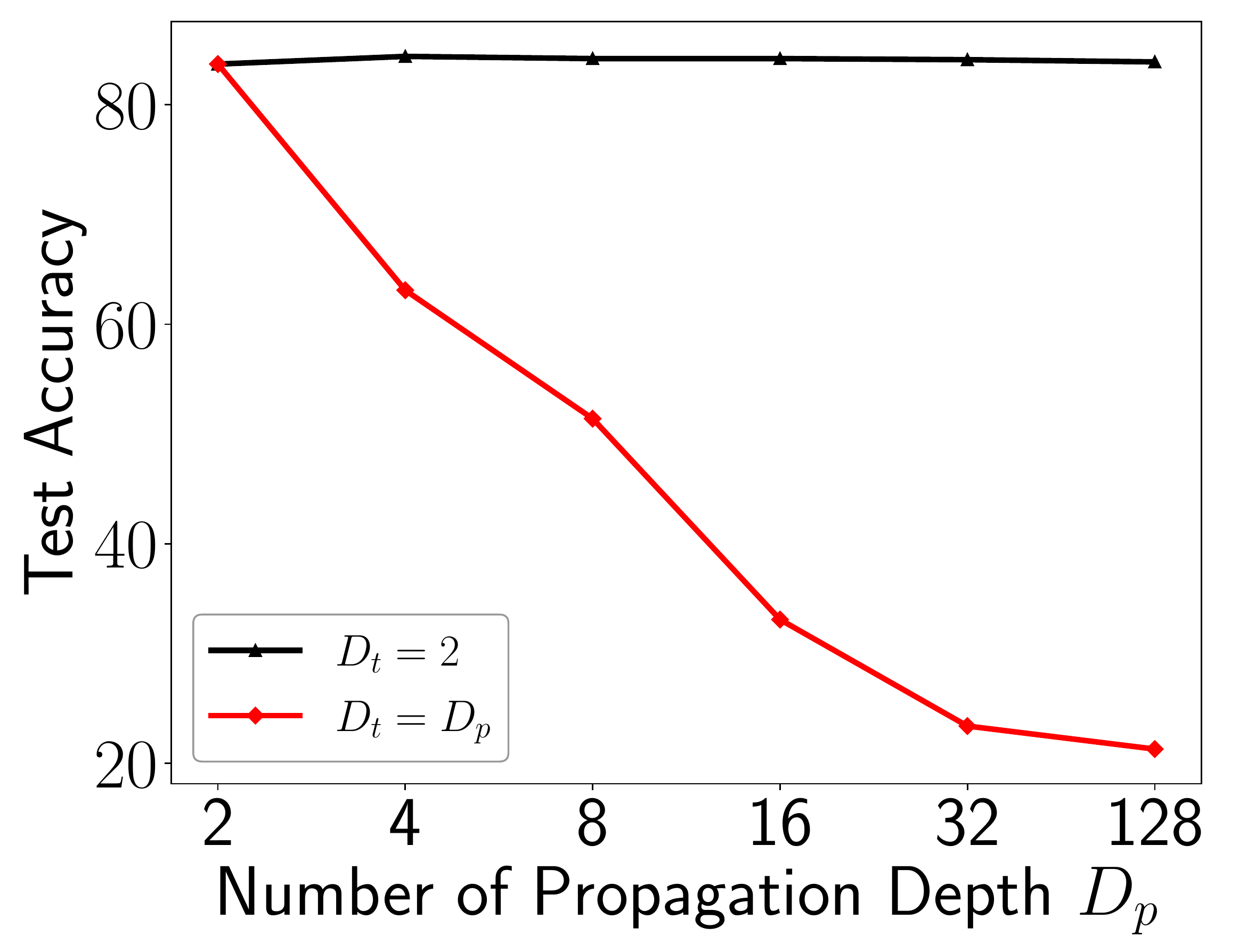}}
\hspace{4mm}
\subfigure[First layer gradient comparison of GCN with different layers on the Cora dataset.]{
\label{fig:gradient}
\includegraphics[width=0.3\textwidth]{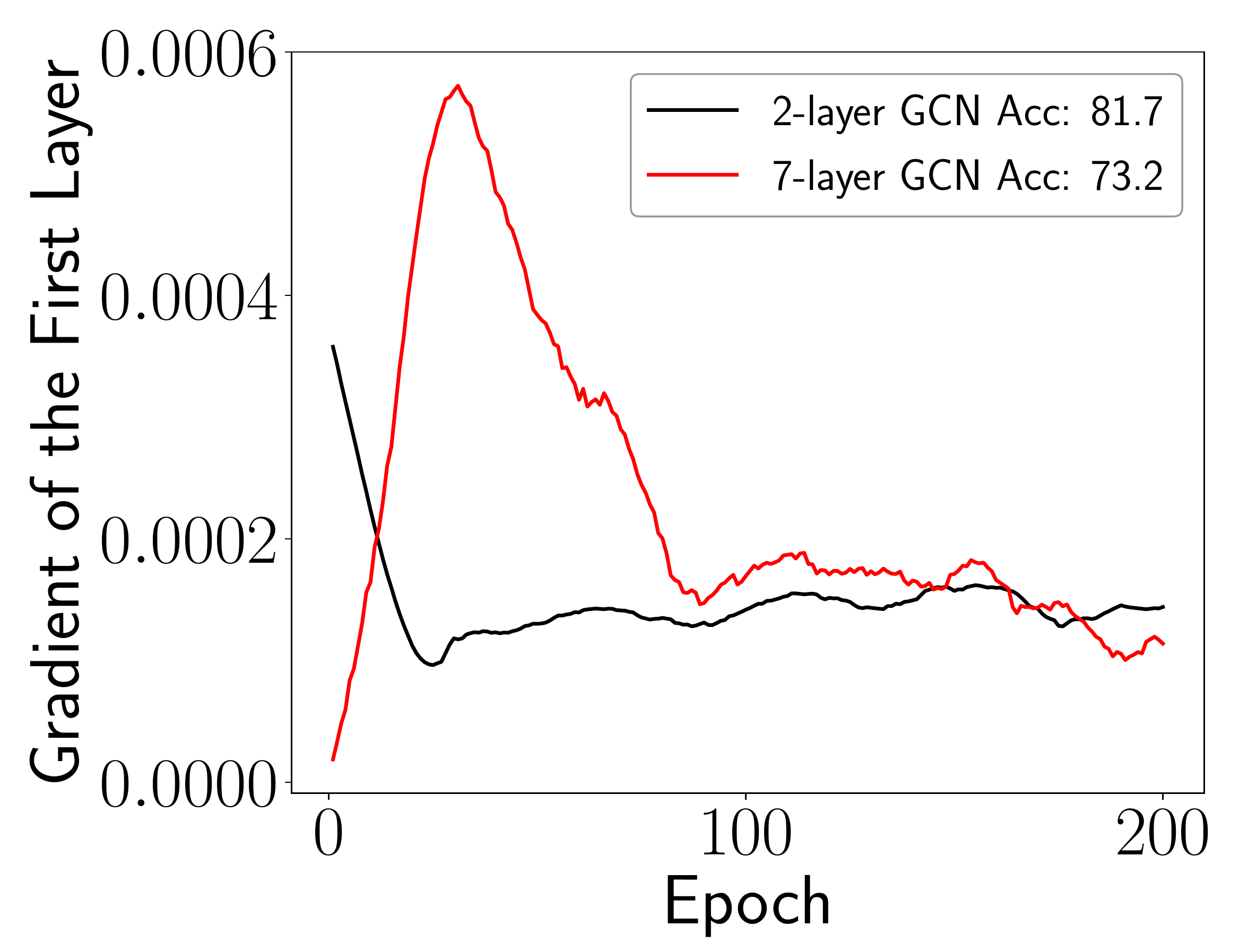}}
\vspace{-3mm}
\caption{Entanglement and gradient vanishing are not the major cause for the performance degradation of deep GNNs.}
\label{fig.three_figs}
\vspace{-2mm}
\end{figure*}

\section{More Misleading Explanations}
\label{app:misconceptions}
\subsection{Entanglement} 
Some recent works~\cite{DBLP:conf/sigir/0001DWLZ020, liu2020towards} argue that the major factor compromising the performance of deep GNNs is the entanglement of the \textbf{P} and \textbf{T} operations in the current graph convolution operation.
However, the evaluation results of ResGCN, DenseGCN, and vanilla GCN on the PubMed dataset when $D_t$ grows in Figure~\ref{fig.res_acc} show that ResGCN and DenseGCN do not encounter significant performance degradation although they are under the entangled design.
Thus, GNNs can go deep even under the entangled design, and the entanglement of the \textbf{P} and \textbf{T} operations may not be the actual limitation of the GNN depth.

What is worth noting is that previous works~\cite{zhu2021simple, liu2020towards}, which disentangle the \textbf{P} and \textbf{T} operations and state that they support deep architectures, only provide experimental results that illustrate they can go deep on $D_p$.
To validate whether they can also go deep on $D_t$, we run DAGNN in two different settings: the first controls $D_t=2$ and increases $D_p$, the second controls $D_t=D_p$ and increases $D_p$.
The test accuracy of DAGNN under these two settings on the PubMed dataset is shown in Figure~\ref{fig:coupled}.

Figure~\ref{fig:coupled} illustrates that DAGNN experiences massive performance decline when the model owns large $D_t$ rather than only large $D_p$.
If the entanglement dominates the performance degradation of deep GNNs, DAGNN should also be able to go deep on $D_t$.
However, the sharp performance decline still exists if we increase $D_t$ along with $D_p$ in DAGNN.
Thus, many recent works that claim they support large GNN depth can only go deep on $D_p$, yet are unable to go deep on $D_t$.

\subsection{Gradient Vanishing}
Gradient vanishing means that the low gradient in the shallow layers makes it hard to train the model weights when the network goes deeper, and it has a domino effect on all of the further weights throughout the network. 

To evaluate whether the gradient vanishing exists in deep GNNs, we perform node classification experiments on the Cora dataset and plot the gradient -- the mean absolute value of the gradient matrix of the first layer in the 2-layer and 7-layer GCN. 
The experimental results are reported in Figure~\ref{fig:gradient}. 

The evaluation results show that the gradients of the 7-layer GCN are as large as the ones of the 2-layer GCN, or even larger in the initial training phases, although the test accuracy of the 7-layer GCN is lower than the one of the 2-layer GCN.
The explanation for the initial gradient rise of the 7-layer GCN might be that the larger model needs more momentum to adjust and then jump out of the suboptimal local minima in the initial model training stage.
Thus, this experiment illustrates that gradient vanishing is not the leading cause for the performance degradation of deep GNNs.

\begin{figure*}[tpb!]
\centering  
\subfigure[Edge Sparsity]{
\includegraphics[width=0.30\textwidth]{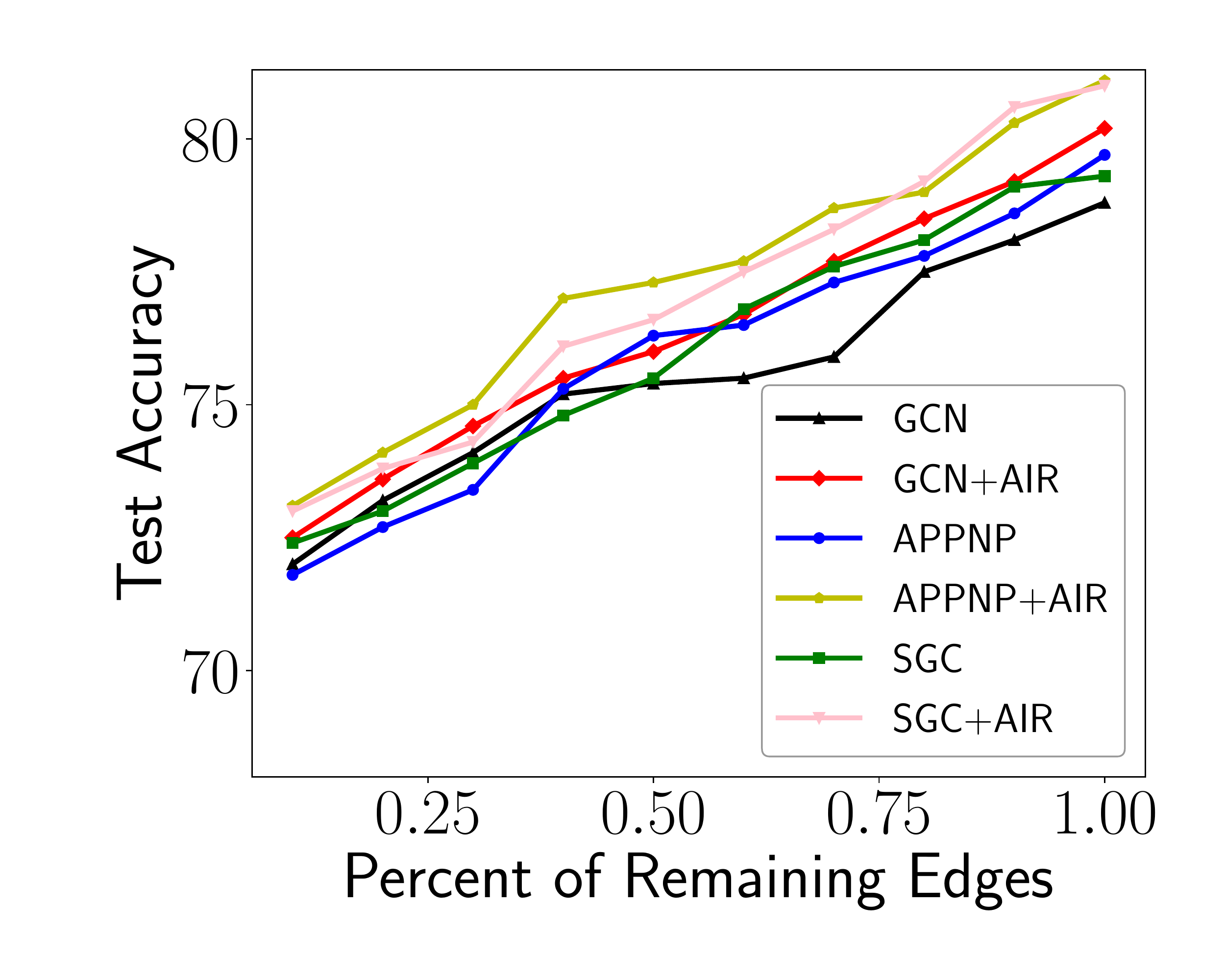}
\label{fig.sparse_edge}
}\hspace{-1mm}
\subfigure[Label Sparsity]{
\includegraphics[width=0.315\textwidth]{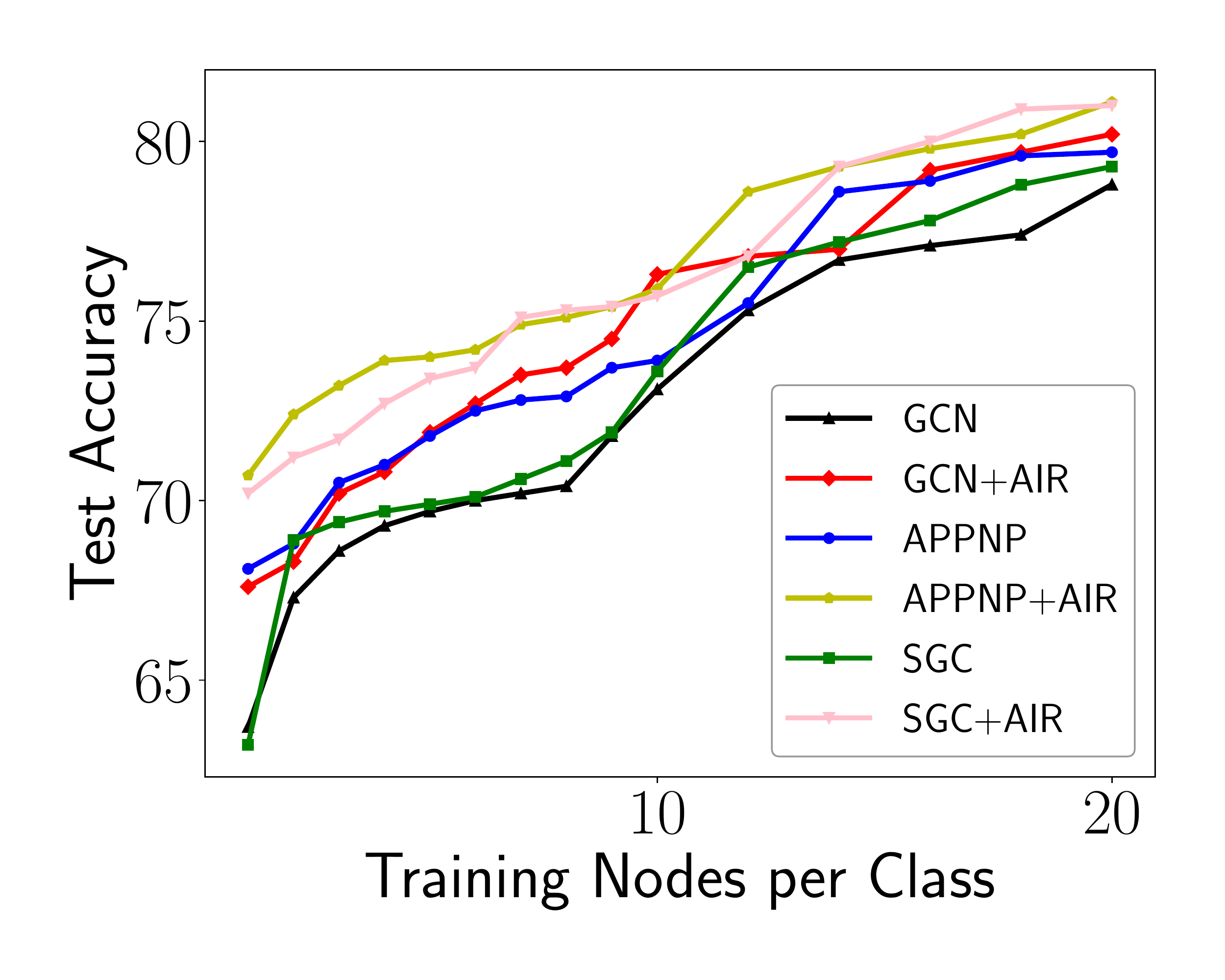}
\label{fig.sparse_label}
}\hspace{-1mm}
\subfigure[Feature Sparsity]{
\includegraphics[width=0.31\textwidth]{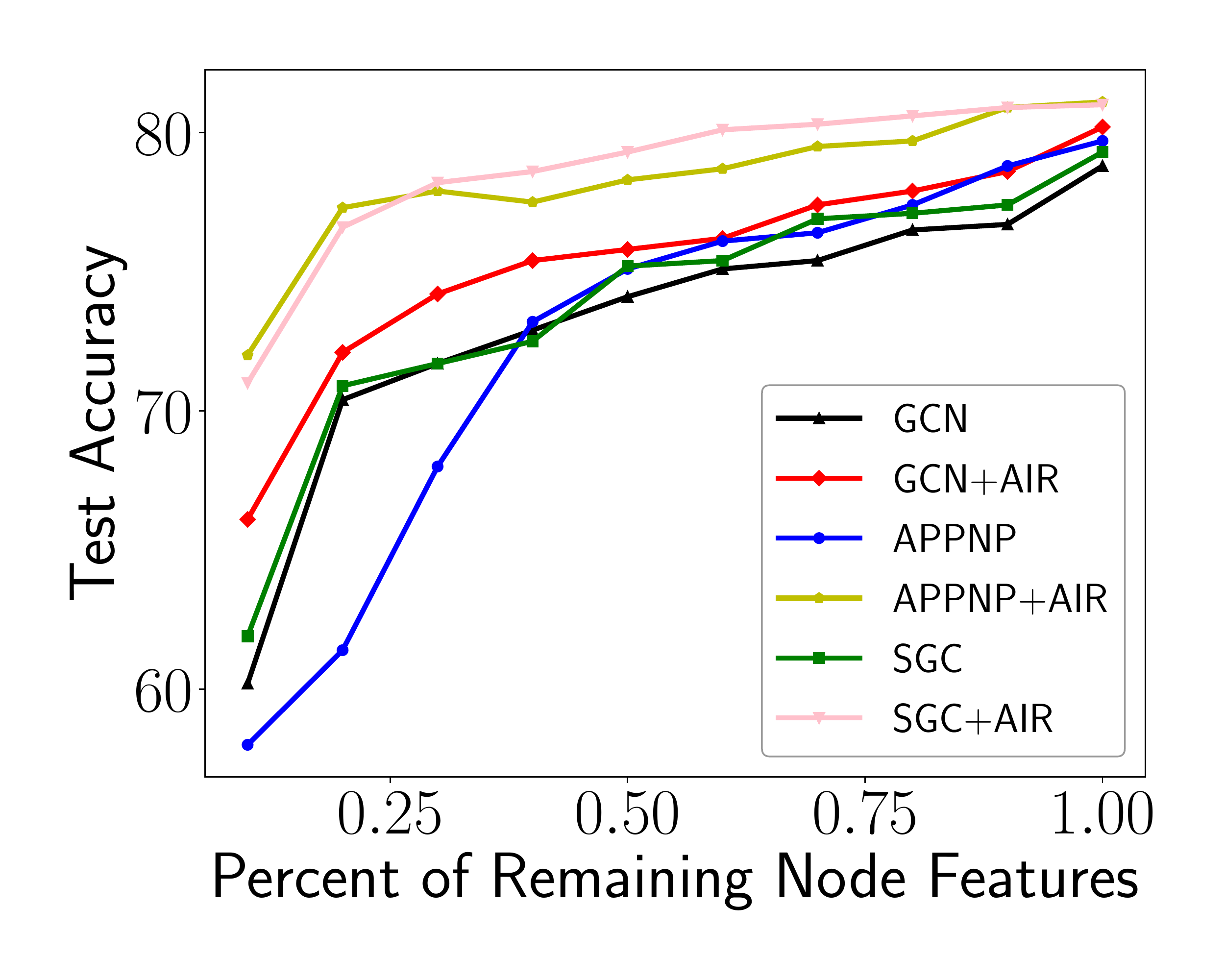}
\label{fig.sparse_feat}
}\hspace{-1mm}
\vspace{-2mm}
\caption{Test accuracy on the PubMed dataset under different levels of feature, edge and label sparsity.}
\label{fig.sparsity}
\vspace{-2mm}
\end{figure*}

\section{Influence of Graph Sparsity on AIR}
\label{sparse}
To simulate extreme sparse scenarios in the real world, we design three sparsity settings on the PubMed dataset to test the performance of our proposed AIR when faced with the edge sparsity, label sparsity, and feature sparsity issues, respectively.

\para{Edge Sparsity.} We randomly remove some edges in the original graph to simulate the edge sparsity situation.
The edges removed from the original graph are fixed across all the compared methods under the same edge remaining rate.
The experimental results in Figure~\ref{fig.sparse_edge} show that all the compared methods perform similarly since the edges play the most crucial role for GNN methods.
However, it can be easily drawn from the results that GCN, APPNP, and SGC all receive considerable performance improvement after being equipped with AIR.

\para{Label Sparsity.} In the label sparsity setting, we vary the training nodes per class from $1$ to $20$ and report the test accuracy of each compared method.
The experimental results in Figure~\ref{fig.sparse_label} show that the test accuracies of all the compared methods increase as the number of training nodes per class ascends.
In the meantime, three GNN methods equipped with AIR all outperform their original version throughout the experiment.

\para{Feature Sparsity.} In a real-world situation, the features of some nodes in the graph might be missing.
We follow the same experimental design in the edge sparsity setting yet remove node features instead of edges.
The results in Figure~\ref{fig.sparse_feat} illustrate that our proposed AIR enables the three GNN baselines great anti-interference abilities when faced with feature sparsity.
For example, the test accuracies of SGC+AIR and APPNP+AIR only drop slightly even there is only $20\%$ node features available.

The above evaluation under three different sparsity settings illustrates that AIR offers excellent help to simple GNN methods.
When adopted on sparse graphs, more \textbf{P} operations in GNNs are always needed since there is more hidden information in the graph, which is probably reachable by long-range connections.
Moreover, more \textbf{T} operations is also preferred since it offers higher expressive power.
Thus, the ability to go deep on $D_p$ and $D_t$ brought by AIR is the main contributor to the great performance improvement of GNN baselines on sparse graphs.

\subsection{Dataset Details}
\label{app:datasets}
Cora, Citeseer, and PubMed are three popular citation network datasets, where nodes stand for research papers, and an edge exists between a node pair if one cites the other.
The raw features of nodes in these three datasets are constructed by counting word frequencies.
The ogbn-arxiv dataset and ogbn-papers100M are also citation networks, yet much bigger than Cora, Citeseer and PubMed.
There are more than 169k and 111M nodes in the ogbn-arxiv dataset and the ogbn-papers100M dataset, respectively.
The ogbn-products dataset is an undirected and unweighted graph representing an Amazon product co-purchasing network.
The details of the adopted six datasets can be found in Table~\ref{datasets}.

\subsection{Experimental Environment}
The experiments are conducted on a server with two Intel(R) Xeon(R) Platinum 8255C CPUs and a Tesla V100 GPU 32GB version.
The operating system of the server is Ubuntu 16.04. 
We use Python 3.6, PyTorch 1.7.1, and CUDA 10.1 for programming and acceleration.

\subsection{Hyperparameter Settings}
\label{app_parameter}
We first provide hyperparameter details on the three citation networks.
For SGC+AIR, $D_t$ is fixed to $2$, and $D_p$ is set to $10$, $15$, and $30$ on the Cora, Citeseer, and PubMed dataset, respectively.
And hidden size is set to $200$ on all three citation networks, while the learning rate is set to $0.1$ on the Cora and Citeseer datasets, and $0.05$ on the PubMed dataset.
For APPNP+AIR, $D_t$, $D_p$ and hidden size is set to $2$, $10$ and $200$ on the three datasets, respectively.
And the learning rate is set to the same values as SGC+AIR.
For GCN+AIR, the number of layers is set to $6$, $4$, $16$ on the Cora, Citeseer, and PubMed datasets, respectively.
The hidden size is set to $32$ on the Cora and PubMed datasets and $16$ on the Citeseer dataset.
And the learning rate is set to $0.01$ on the Cora and Citeseer datasets and $0.1$ on the PubMed dataset. 

The hyperparameter settings on the three ogbn datasets are as follows:
For SGC+AIR, $D_t$ is set to $6$ on the ogbn-arxiv and ogbn-papers100M datasets, $2$ on the ogbn-products dataset.
$D_p$ is set to $5$ on the ogbn-arxiv and ogbn-products datasets, $15$ on the ogbn-papers100M dataset.
The hidden size and learning rate are set to $0.001$ and $1024$ on all three datasets.
On the ogbn-arxiv dataset, $D_p$ and hidden size are set to $16$ and $256$ for APPNP+AIR and GCN+AIR.
$D_t$ is set to $2$ and $16$ for APPNP+AIR and GCN+AIR, respectively.
The learning rate is set to $0.005$ and $0.001$ for APPNP+AIR and GCN+AIR, respectively.

Other hyperparameters are tuned with the toolkit OpenBox~\cite{li2021openbox} or follow the settings in their original paper.
The source code can be found in Anonymous Github (\textit{\blue{\url{https://github.com/PKU-DAIR/AIR}}}).
Please refer to ``README.md'' in the Github repository for more reproduction details.

\end{document}